\documentclass[10pt,journal]{IEEEtran}

\emergencystretch=\maxdimen
\usepackage{times}
\usepackage{epsfig}
\usepackage{graphicx}
\usepackage{amsmath}
\usepackage{amssymb}

\usepackage[pagebackref=true,breaklinks=true,letterpaper=true,colorlinks=true,bookmarks=true,bookmarksdepth=3]{hyperref}

\usepackage{floatrow}

\usepackage{color}
\usepackage{epsfig}
\usepackage{epstopdf}
\usepackage{caption}
\usepackage{cellspace}
\usepackage{subfloat}
\usepackage{subcaption}
\usepackage{lineno}
\usepackage{float}
\usepackage{units}
\usepackage{bm}
\usepackage{enumerate}
\usepackage{multirow}
\usepackage{rotating}
\usepackage[hang,flushmargin]{footmisc}
\usepackage{mdwlist}
\usepackage{multirow}
\usepackage{lipsum}
\usepackage{textcomp}
\usepackage[table,xcdraw,dvipsnames]{xcolor}
\usepackage{tablefootnote}
\usepackage{threeparttable}

\usepackage{algorithm}
\usepackage[noend]{algpseudocode}
\usepackage{pgffor}

\usepackage{enumitem,kantlipsum}
\usepackage{url}
\usepackage{varwidth}

\usepackage[table,xcdraw]{xcolor}
\usepackage{pagecolor}
\usepackage{makecell}

\usepackage{xr}

\hypersetup{
	colorlinks=true,       % false: boxed links; true: colored links
	linkcolor=blue,          % color of internal links (change box color with linkbordercolor)
	citecolor=red,        % color of links to bibliography
	urlcolor=magenta           % color of external links
}

\algnewcommand\algorithmicforeach{\textbf{for each}}
\algdef{S}[FOR]{ForEach}[1]{\algorithmicforeach\ #1\ \algorithmicdo}
\algnewcommand\And{\textbf{and}}

\newcolumntype{M}{>{\centering\arraybackslash}m{\dimexpr.10\linewidth-2\tabcolsep}}

\begin{document}
%\linenumbers
%%%%%%%%% TITLE
%\title{River Ice Concentration Analysis with Deep Learning}
\title{Deep MDP: A Modular Framework for Multi-Object Tracking}
%\author{Abhineet Singh \hspace{1cm} Nilanjan Ray \\
%	University of Alberta\\
%	{\tt\small asingh1,nray1@ualberta.ca}
%}
\author{Abhineet Singh\\
	University of Alberta\\
	{\tt\small asingh1@ualberta.ca}
}

\maketitle

%\linenumbers

\begin{abstract}
This paper presents a fast and modular framework for Multi-Object Tracking (MOT) based on the Markov descision process (MDP) tracking-by-detection paradigm.
It is
%termed Deep MDP and
designed to allow its various functional components to be replaced by custom-designed alternatives to suit a given application.
An interactive GUI with integrated object detection, segmentation, MOT and semi-automated labeling is also provided to help make it easier to get started with this framework.
Though not breaking new ground in terms of performance, Deep MDP has a large code-base that should be useful for the community to try out new ideas
%in this domain
or simply to have an easy-to-use and easy-to-adapt system for any MOT application.
%We demonstrate its deployment in the domain of cell tracking.
Deep MDP is available at ~\url{https://github.com/abhineet123/deep_mdp}.
%and segmentation.
\end{abstract}

%%%%%%%%% BODY TEXT
\section{Introduction}
\label{sec_intro}
MOT is the problem of detecting and tracking each object that enter the scene in a video to construct its trajectory.
It is an important mid-level task in computer vision
%that builds on lower level tasks like object classification, detection and segmentation and is used for many high level tasks like
with wide ranging applications including 
autonomous driving,
road traffic analysis,
%for highway and intersection design,
video surveillance,
activity recognition,
medical imaging,
human-computer interaction and
virtual reality. 
Most MOT methods employ the tracking-by-detection paradigm \cite{mot_review17} where an object detector \cite{det_review_ijcv20} first finds likely objects
in each video frame
and the tracker then
associates objects across frames
%uses an association algorithm \cite{assignment_review18}
to assign a unique ID to all detections corresponding to each object instance.

\section{MDP}
\label{mdp}
\subsection{Overview}
\label{mdp-overview}
\begin{figure}[t]
	\centering
		\caption{
			\small{
				State transition diagram for MDP borrowed from \cite{Xiang15_mdp}.
				\textit{Active} actions $ a_1 $ and $ a_2 $ respectively correspond to TP and FP detections.
				\textit{Tracked} actions $ a_3 $ and $ a_4 $ respectively correspond to an object remaining trackable (e.g. clearly visible) and becoming untrackable (e.g. occluded) in the current frame.
				\textit{Lost} actions $ a_6 $ and $ a_5 $ are analogous to  $ a_3 $ and $ a_4 $  while $ a_7 $ corresponds to an object that has permanently left the scene.
				Though not shown here or mentioned in the paper, the MDP code also has a transition from \textit{tracked} to \textit{inactive} which is made using the same sort of heuristics as $ a_7 $.						
			}
		}
		\label{fig-mdp_state_transition}
	{
	\includegraphics[width=\textwidth]{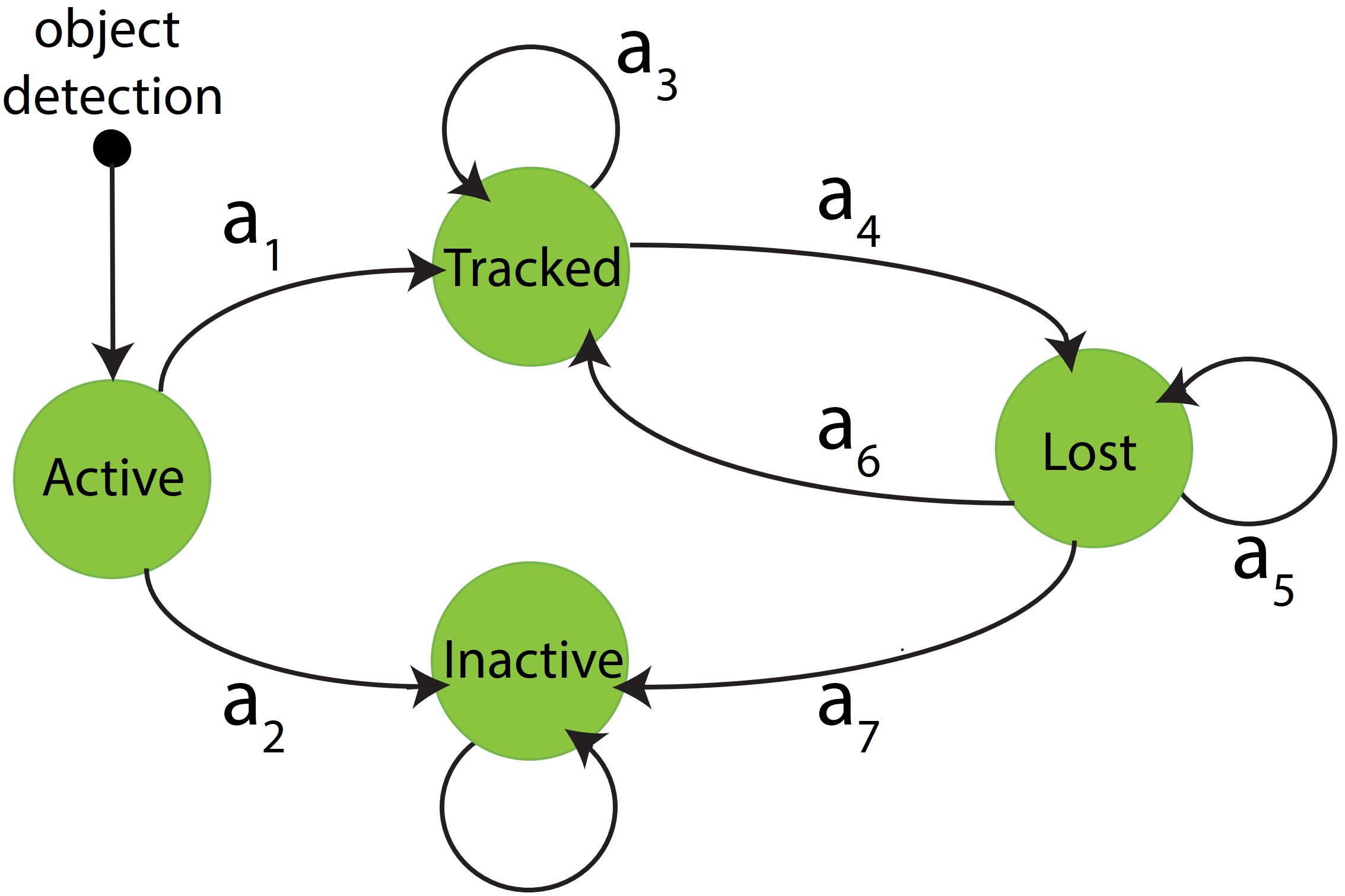}
	}
	
	%	\label{fig-mdp_state_transition}
\end{figure}

The main idea of the MDP framework
%for MOT
is to conceptualize a tracked object as a Markov Decision Process so that it is in one of a certain number of discrete states in each frame and transitions between these states according to well-defined and potentially learnable rules called policies.
It was originally introduced in \cite{Xiang15_mdp} along with an updated version in \cite{Sadeghian17_mdp2} although similar ideas has been used for target management in several other trackers \cite{Chu2017_spatio_temporal_mdp,Zhu2018_dman_mdp,Wan21_end_to_end_mot}.
For the remainder of this paper, MDP refers only to the original version \cite{Xiang15_mdp}
%on which I have done most of my work so far.
%An object in this variant can be in one of
that defines four states (Fig. \ref{fig-mdp_state_transition}):
%\begin{itemize}[noitemsep,topsep=0pt]
\begin{itemize}[left=0pt,topsep=0pt,noitemsep,label=\textendash]
	\item a potential new object detected for the first time moves into the \textbf{\textit{active}} state
	\item an existing object that is successfully tracked in the current frame (e.g. if it is clearly visible) moves into or remains in the \textbf{\textit{tracked}} state
	\item an existing object that can not be successfully tracked in the current frame (e.g. if it is occluded) moves into or remains in the  \textbf{\textit{lost}} state
	\item an object that has left the scene permanently moves into the \textbf{\textit{inactive}} state
\end{itemize}
\textit{Active}, \textit{lost} and \textit{tracked} have their own transition policies while \textit{inactive} is a sink state for discarded objects and thus has no transitions.
\textit{Active} policy determines if a new detection unmatched with any existing target is a real object to start tracking or an FP to be discarded and makes respective transitions to \textit{tracked} and \textit{inactive}.
MDP makes this decision using a binary SVM classifier \cite{Boser92_svm} trained with handcrafted features based on shape, size and position of boxes.
%under the assumption that new objects enter the scene at a specific positions in the frame and have certain types of shapes and sizes.
\textit{Tracked} policy determines whether or not a target that was successfully tracked in the previous frame in also trackable in the current frame, with respective transitions to \textit{tracked} and \textit{lost}.
MDP makes this decision purely with heuristics based on the success in tracking the object
%from its previous location
through forward-backward LK tracking \cite{Bouguet00_pyr_lk,Kalal2012_tld} and
%the overlap of its predicted location
%its overlap with any
the presence of a matching
detection in the current frame.
\textit{Lost} policy decides whether the target is trackable in the current frame, still untrackable or has permanently left the scene, with respective transitions to \textit{tracked}, \textit{lost} and \textit{inactive}.
The \textit{lost} / \textit{tracked} transition decision is made by a binary SVM classifier
that predicts whether the target matches a detection using
%trained on
handcrafted features respresnting the similarity of their appearance and shapes as well as the success of LK tracking from each to the other.
Transition to \textit{inactive} is done by thresholding on the
%proximity of its location to the edge of the image
overlap of its bounding box with the image
and
%the duration for which it has
%continuously
%been in the \textit{lost} state.
how long it has been in the \textit{lost} state.

In addition to these policies, MDP employs several more target-level heuristics including a set of templates to represent the changing object appearance and a constant velocity motion model
to predict its location
in the current frame
There are also many global heuristics involved in inter-target processing like detection filtering, target conflict resolution, recovering
recently
\textit{lost} targets and
%postprocessing
discarding short
trajectories (Alg. \ref{alg-mdp_testing}).

\subsection{Pairwise Tracking of Templates}
\label{mdp-lk}
\begin{table*}[!htbp]
	\centering
	\caption{
		\small{
			Impact of the number of templates on tracking performance.
			Results were generated using original MDP on DETRAC dataset but similar trends hold with deep MDP as well as over other datasets.
		}
	}
	\begin{tabular}{|c|ccc|cc|}
		\hline
		\multirow{2}{*}{\makecell{\textbf{no. of}\\\textbf{templates}}} & \multicolumn{3}{c|}{\textbf{higher is better}}                                                      & \multicolumn{2}{c|}{\textbf{lower is better}}       \\ \cline{2-6} 
		& \multicolumn{1}{c|}{\textbf{MOTA (\%)}} & \multicolumn{1}{c|}{\textbf{MOTP (\%)}} & \textbf{MT(\%)} & \multicolumn{1}{c|}{\textbf{ML(\%)}} & \textbf{IDS} \\ \hline
		\textbf{2}                                    & \multicolumn{1}{c|}{84.81}             & \multicolumn{1}{c|}{54.84}            & 91.52       & \multicolumn{1}{c|}{2.45}        & 2592         \\ \hline
		\textbf{5}                                    & \multicolumn{1}{c|}{84.68}             & \multicolumn{1}{c|}{55.98}            & 91.74       & \multicolumn{1}{c|}{2.23}        & 2641         \\ \hline
		\textbf{10}                                   & \multicolumn{1}{c|}{84.13}              & \multicolumn{1}{c|}{56.17}            & 92.52       & \multicolumn{1}{c|}{1.78}        & 2856         \\ \hline
	\end{tabular}
	\label{tab-templates}
\end{table*}
MDP uses tracking more as a way to measure the similarity between two patches representing prospective objects in different frames than to actually track objects over time.
Given the object locations in two possibly non-consecutive frames, an ROI centered on the object is extracted from each frame,
with
a certain amount of its background included for context,
and then scaled anisotropically
%subjected to a non-aspect ratio preserving scaling
to a fixed size.
The default parameters, tuned for tall boxes corresponding to pedestrians, result in an object size of $ 60 \times 45 $ with a $ 120 \times 45 $ border for an ROI size of $ 300 \times 135 $.
Forward-backward LK tracking \cite{Kalal2012_tld} is then applied on the  $ 60 \times 45 $ boxes between the two ROIs and its success is used as a proxy for how similar the two objects are.
This kind of discontinuous ROI-to-ROI tracking works with optical flow methods like LK but it is not suitable for normal long-term trackers.

One of the two ROIs being tracked corresponds to either the predicted location (in \textit{tracked}) or a detection (in \textit{lost}), both from the current frame.
The other ROI is one of several templates, which are ROIs extracted from heuristically-chosen keyframes from the
%object's history.
history of the object.
%The other one corresponds to a predicted location in \textit{tracked} and a detection in \textit{lost}, both in the current frame.
%\subsubsection{Template summarization}
Since each template is tracked independently,
%with each detection or predicted location,
this operation not only consumes most of the runtime but also requires additional heuristics to summarize the tracking performance over all the templates
in order
%so as
to
generate a single representative feature that can be passed to the classifier.
MDP uses a heuristically-selected anchor template
%that supposedly best represents the object
to generate this summary feature,
thus rendering the expensive template-tracking operation virtually ineffective (Table \ref{tab-templates}).
%\subsubsection{Detection Override}
Another heuristic that undermines not only the tracker but the classifier itself is that both \textit{tracked} and \textit{lost} policy decisions depend on the presence of a matching detection 
and it is the latter that gets priority since a missing detection
always
results in a negative policy decision irrespective of the classification result.
All the costly tracking and classification therefore amounts to little more than an additional verification step to confirm that a detection near the expected location of the object actually corresponds to that object.

\subsection{Training}
\label{mdp-training}
Training the MDP system
%This
involves training the two SVM classifiers for \textit{active} and \textit{lost} policies.
\textit{Active} SVM is trained in batch on samples generated by applying two-level thresholding on the intersection-over-union (IOU) of detections with the ground truth (GT) that classifies the former into false positive (FP), true positive (TP) or unknown.
FP and TP detections respectively become negative and positive training samples while those classified as unknown are ignored.
On the other hand, \textit{lost} SVM is trained incrementally by tracking one trajectory at a time (Alg. \ref{alg-mdp_training})
using
%while making policy decisions with 
an untrained or partially trained model
for policy decisions,
collecting a new sample on each incorrect decision and retraining the model with the new sample added to existing training set.
This is a major limitation of this framework not only because trajectory-level training does not allow incorporating inter-target interactions (Alg. \ref{alg-mdp_testing}) that play a major role at inference (Table \ref{tab-global_heuristics}) but also with respect to adaptation for deep learning since training deep networks incrementally leads to overfitting and prevents convergence to a good optimum.
Another limitation of \textit{lost} training is that samples can only be extracted from frames where a detection corresponding to the GT is present so that there is no way of learning from FNs.

\subsection{Modular Implementation}
\label{mdp-modular}
I started by adapting the MATLAB implementation of MDP \cite{mdp_github} into a fast, parallelized and modular Python version \cite{deep_mdp_github}
for real-time vehicle and pedestrian tracking in road traffic videos of intersections
captured
from pole-mounted and UAV cameras.
I also integrated it into an interactive GUI \cite{mdp_labeling_tool_github} for object detection, MOT and semi-automated labeling.
One of my main design objectives was to make it modular so that existing components could be easily swapped with alternatives from deep learning.

\begin{figure}[t]
	\includegraphics[width=\textwidth]{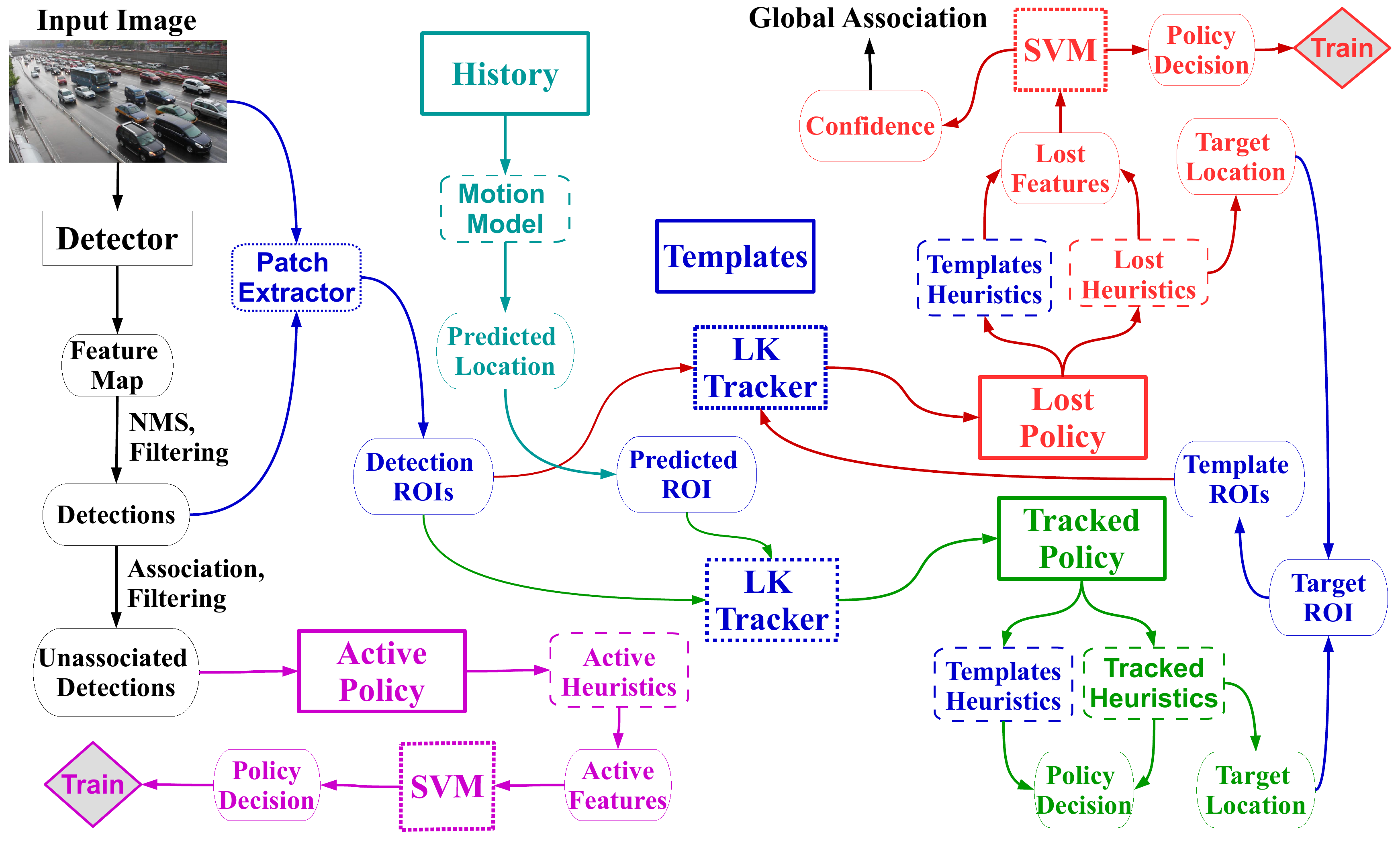}
	\includegraphics[width=\textwidth]{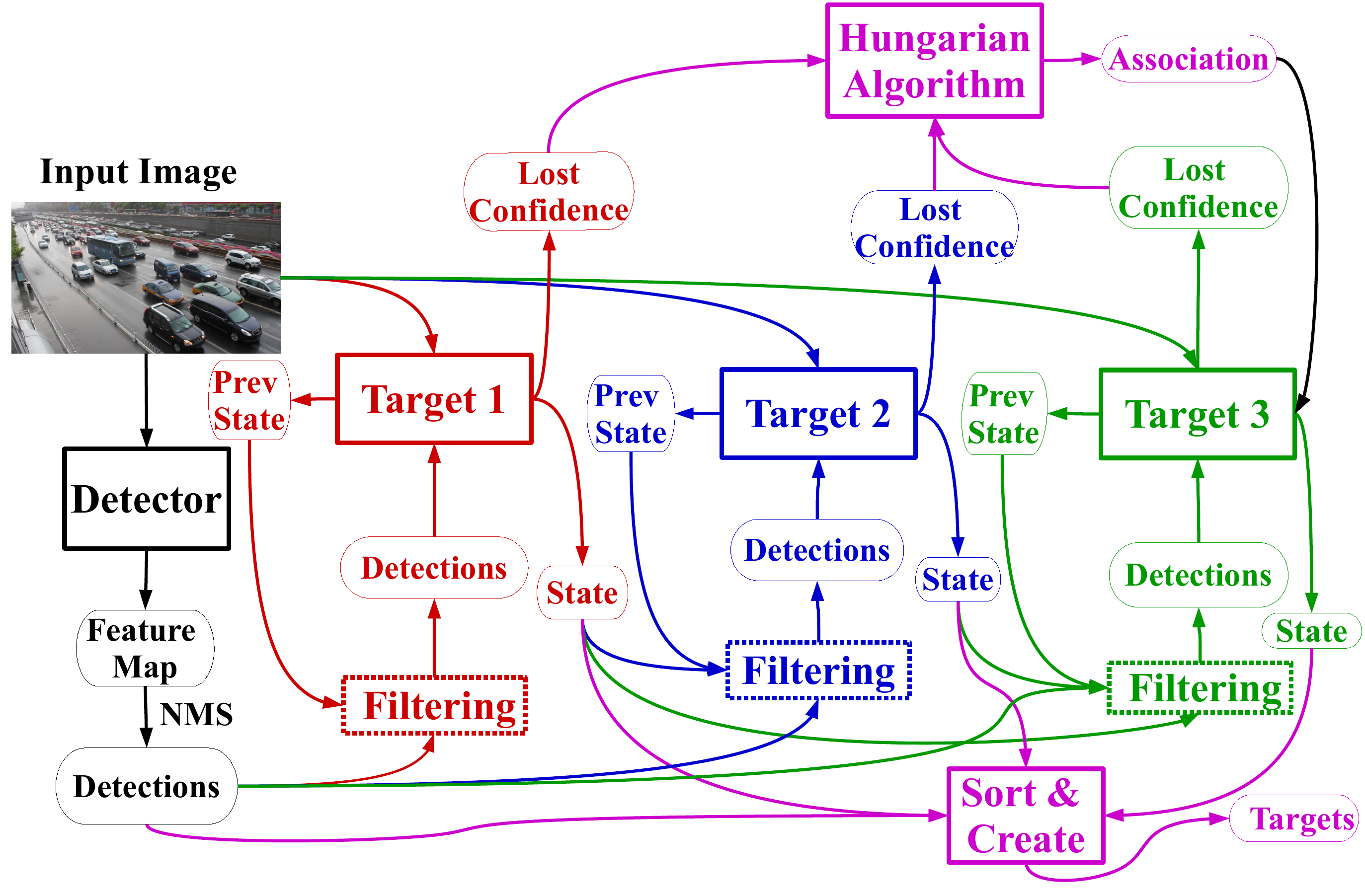}
	\caption{
		\small{
			MDP modular implementation block diagrams for (top) target-level and (bottom) global processing.
			Left: Colors to modules - blue: \texttt{Templates}, cyan: \texttt{History}, red: \texttt{Lost}, green: \texttt{Tracked}, magenta: \texttt{Active}.
			Lines connecting the various submodules of \texttt{Templates} are not shown for clarity.			
			Right: This shows the inference process for each frame (Alg. \ref{alg-mdp_testing}). Lines connecting association box to targets 1 and 2 are not shown for clarity.
			%			Full page version of these diagrams in 
		}			
	}
	\label{fig-mdp_block_diag}
\end{figure}

There are five main target-level modules in this implementation (Fig. \ref{fig-mdp_block_diag}) -  \texttt{Templates}, \texttt{History}, \texttt{Lost}, \texttt{Tracked} and \texttt{Active}.
The latter three correspond to the respective state policies while \texttt{Templates} implements all the functionality that is shared between \texttt{Tracked} and \texttt{Lost}.
This includes creating and updating the list of templates representing the object appearance over time, tracking, ROI extraction and heuristics-based feature generation for classification.
There are two main global modules - \texttt{Trainer} that trains policies (Alg. \ref{alg-mdp_training}, Sec. \ref{mdp-training}) and \texttt{Tester} that performs inference (Alg. \ref{alg-mdp_testing}).
Feature extraction, classification and tracking are also implemented as independent modules but used as submodules within the main modules.

\begin{table*}[!htbp]
	\centering
	\caption{
		\small{
			Impact of global heuristics in the original MDP \texttt{Tester} on DETRAC tracking performance.
			These include target sorting, detection filtering, reconnecting recently \textit{lost} targets and resolving conflicts between \textit{tracked} targets.
			These results also demostrate the inconsistency that often exists between the various MOT metrics in that better performance over some metrics is accompanied by worse performance over others.
			% Results were generated using original MDP on sequences 30-50 of DETRAC.
		}
	}
	\begin{tabular}{|c|ccc|cc|}
		\hline
		\multirow{2}{*}{\textbf{config}} & \multicolumn{3}{c|}{\textbf{higher is better}}                                                      & \multicolumn{2}{c|}{\textbf{lower is better}}       \\ \cline{2-6} 
		& \multicolumn{1}{c|}{\textbf{MOTA (\%)}} & \multicolumn{1}{c|}{\textbf{MOTP (\%)}} & \textbf{MT(\%)} & \multicolumn{1}{c|}{\textbf{ML(\%)}} & \textbf{IDS} \\ \hline
		\textbf{heuristics   on}         & \multicolumn{1}{c|}{84.812}             & \multicolumn{1}{c|}{54.844}            & 91.524          & \multicolumn{1}{c|}{2.454}           & 2592         \\ \hline
		\textbf{heuristics   off}        & \multicolumn{1}{c|}{67.629}             & \multicolumn{1}{c|}{58.834}            & 97.089          & \multicolumn{1}{c|}{0.960}            & 9435         \\ \hline
	\end{tabular}
	\label{tab-global_heuristics}
\end{table*}

\begin{algorithm}[!htbp]
	%\begin{algorithm}[t]
	\caption{
		MDP Training in \texttt{Trainer} module
		%		filterTrajectories returns
	}				
	\label{alg-mdp_training}
	\footnotesize
	%	\scriptsize	
	\begin{algorithmic}[1]
		\State \textit{max\_passes}, \textit{max\_trials}, \textit{max\_iters} $\gets$ parameters
		\State \textit{trajectories}, \textit{detections} $\gets$ all trajectories and detections in training set		
		%		\State \textit{trajectories} $\gets$ all trajectories in training set	
		\State \textit{trajectories} $\gets$ filterTrajectories(\textit{trajectories}, \textit{detections})
		\Comment{remove trajectories not suitable for \texttt{lost} training}
		\State mark all \textit{trajectories} as \texttt{trainable}
		%		\State \textit{t}.\textit{trainable} $\gets$ \textbf{true} $ \forall $ \textit{t} $ \in $ \textit{trajectories}
		\State \textit{pass} $\gets$ 0
		\State \textit{train\_samples} $\gets$ $ \emptyset $
		\For{\textit{iter} $\gets$ 1, \textit{max\_iters}}
		\State mark all \texttt{trainable} \textit{trajectories} as \texttt{not done}
		%			\State \textit{t}.\textit{done} $\gets$ \textbf{false} $ \forall $ \textit{t} $ \in $ \textit{trajectories} $ | $ \textit{t}.\textit{trainable} = \textbf{true}
		%			\State \textit{t}.\textit{trials} $\gets$ 0 $ \forall $ \textit{t} $ \in $ \textit{trajectories}
		\State set \texttt{trials} to 0 for all \textit{trajectories}
		%\ForEach{\textit{trajectory} $ \in $  \textit{train\_trajectories}}	
		%\State \textit{trajectory}.\textit{done} $\gets$ \textbf{false}
		%\State \textit{trajectory}.\textit{trials} $\gets$ 0				
		%\EndFor
		\If{
			%				\textit{t}.\textit{trainable} =  \textbf{false} $\forall $ \textit{t} $ \in $ \textit{trajectories} 
			all \textit{trajectories} marked as \texttt{untrainable}
		}
		\State \textbf{break}			
		\ElsIf{
			%				\textit{t}.\textit{done} =  \textbf{true} $\forall $ \textit{t} $ \in $ \textit{trajectories} 
			all \textit{trajectories} marked as \texttt{done}
		}
		\State \textit{pass} $\gets$ \textit{pass} + 1
		\If{\textit{pass} $ > $ \textit{max\_passes}}
		\State \textbf{break}
		\EndIf
		\Else
		%				\State \textit{trajectory} $\gets$ find a \textit{t} $ \in $  \textit{trajectories} $ | $ \textit{t}.\textit{done} =  \textbf{false}
		\State \textit{trajectory} $\gets$ find the next trajectory marked as \texttt{not done}
		\EndIf
		%\ForEach{\textit{trajectory} $ \in $  \textit{train\_trajectories}}			
		%\State \textit{target} $\gets$ initialize new target in \textit{frames}[0]
		\State \textit{frames} $\gets$ all frames in \textit{trajectory}
		\State \textit{failures} $\gets$ 0
		\ForEach{\textit{frame} $ \in $ \textit{frames}}
		\State \textit{dets} $\gets$ all detections in \textit{frame}
		%\State \textit{gts} $\gets$ all annotations in \textit{frame}
		%\State \textit{success}  $\gets$ \textit{target}.update(\textit{frame}, \textit{dets})
		\State \textit{success}  $\gets$ track this object in \textit{frame} using \textit{dets}
		\Comment{\textit{success} indicates correctness of policy decision}
		%and check policy decision with \textit{gts}
		\If{\textbf{not} \textit{success} \textbf{and} object is in \texttt{lost} state}
		\State	\textit{new\_sample} $\gets$ sample corresponding to incorrect policy decision
		\State \textit{train\_samples} $\gets$ \textit{train\_samples} $ \cup $ \textit{new\_sample} 
		\State train lost policy with \textit{train\_samples}
		\State \textit{failures} $\gets$ \textit{failures} + 1
		\EndIf
		\EndFor
		\If{\textit{failures} $ = $ 0}
		%\State \textit{trajectory}.\textit{done} $\gets$ \textbf{true}		
		\State mark \textit{trajectory}	as \texttt{done}
		%				\State \textit{trajectory}.\textit{done} $ \gets $  \textbf{true}
		\Comment{trajectory successfully tracked in entirety}
		\Else
		%\State \textit{trajectory}.\textit{trials} $\gets$ \textit{trajectory}.\textit{trials} + 1		
		\State increment \textit{trajectory} \texttt{trials} by 1 		
		%				\State \textit{trajectory}.\textit{trials}	 $ \gets $ \textit{trajectory}.\textit{trials} + 1
		\If{
			\textit{trajectory} \texttt{trials} $ > $ \textit{max\_trials}
			%					\textit{trajectory}.\textit{trials} $ > $ \textit{max\_trials}
		}
		% \State \textit{trajectory}.\textit{done} $\gets$ \textbf{true}	
		\State mark \textit{trajectory} as \texttt{untrainable}
		%					\State \textit{trajectory}.\textit{trainable} $ \gets $  \textbf{false}
		\Comment{trajectory is too difficult to track}
		
		\State mark \textit{trajectory} as \texttt{done}
		%					\State \textit{trajectory}.\textit{done} $ \gets $  \textbf{true}
		\EndIf
		\EndIf
		% \EndFor		
		\EndFor
		
		\State
		
		\Function {filterTrajectories}{\textit{trajectories}, \textit{detections}}

		\State truncate each trajectory to start in the first frame where:
		\State \hskip\algorithmicindent 1. object has at least one detection with IOU $ > $ 0.5
		\State \hskip\algorithmicindent 2. object does not have IOU $ > $ 0 with any other annotation
		\State \hskip\algorithmicindent 3. IOU of object with image $ > $ 0.95
		\State remove trajectories that have no such frame	
		\EndFunction
	\end{algorithmic}
\end{algorithm}

\begin{algorithm}[h]
	\centering
	%\begin{algorithm}[t]
	\caption{
		MDP Inference in \texttt{Tester} module
	}				
	\label{alg-mdp_testing}
	%	\scriptsize	
	\footnotesize
	\begin{algorithmic}[1]	
		\State \textit{hungarian} $\gets$ parameter
		\State \textit{frames} $\gets$ all frames in video
		\State \textit{trajectories} $\gets \emptyset $
		\State \textit{targets} $\gets \emptyset $
		\ForEach{\textit{frame} $ \in $ \textit{frames}}
		\State \textit{dets} $\gets$ all detections in \textit{frame}
		\State \textit{sorted} $\gets$ sortTargets(\textit{targets})
		\ForEach{\textit{target} $ \in $ \textit{sorted}}
		\State \textit{prev\_targets} $\gets$ targets processed so far
		\If{\textit{target}.\textit{state} = \texttt{lost}}
		\State \textit{filtered} $\gets$ filterDetections(\textit{dets}, \textit{prev\_targets})
		\Else
		\State \textit{filtered} $\gets$ \textit{dets}
		\EndIf	
		%				\State\textit{target}.update(\textit{frame}, \textit{filtered})
		\State track \textit{target} in \textit{frame} using \textit{filtered}
		\State\textit{target}	 $\gets$ apply heuristics to check if \textit{target} left the scene
		\If{\textit{target}.\textit{state} = \texttt{inactive}}
		\State \textit{trajectories} $\gets$ \textit{trajectories} $ \cup $ \textit{target}.\textit{trajectory}
		\State \textit{targets} $\gets $ \textit{targets} $ \setminus $ \textit{target}
		\EndIf
		\If{
			\textit{target}.\textit{state} = \texttt{lost}
			\textbf{and}
			\textit{target}.\textit{prev\_state} = \texttt{tracked}
		}
		\Comment{try to reconnect recently lost target}
		\State \textit{filtered} $\gets$ filterDetections(\textit{dets}, \textit{prev\_targets})							
		%					\State \textit{target}.update(\textit{frame},\textit{filtered})		
		\State track \textit{target} in \textit{frame} using \textit{filtered}				
		\EndIf				
		\EndFor
		\If{\textit{hungarian}}
		\Comment{otherwise each target associates to highest-probability detection}			
		\State \textit{cost\_matrix} $\gets$ accumulate classifier probabilities from \texttt{lost} targets
		\State \textit{association\_matrix} $\gets$ HungarianAlgorithm(\textit{cost\_matrix})
		\State \textit{targets} $\gets $ associate \textit{targets} to \textit{dets} using \textit{association\_matrix}
		\EndIf
		\State \textit{unassociated\_dets} $\gets$ filterDetections(\textit{dets}, \textit{targets})
		\State \textit{new\_targets} $\gets $ apply active policy to \textit{unassociated\_dets}
		\State \textit{targets} $\gets $ \textit{targets} $ \cup $ \textit{new\_targets}
		\State \textit{targets} $\gets $ resolveTargetConflicts( \textit{targets} )
		\EndFor
		\State \textit{trajectories} $\gets $ remove trajectories $ <  $ 5 frames long
		\State
		\Function {sortTargets}{\textit{targets}}
		\State divide \textit{targets} into two subsets with \texttt{tracked} streaks of  $ > 10 $ and $ \leq 10 $ frames 
		\State sort each subset to place \texttt{tracked} targets before \texttt{lost} targets
		\State
		%			\textit{sorted\_targets} $\gets $
		join the two sorted subsets
		%			\State \textbf{return}  \textit{sorted\_targets}
		\EndFunction		
		\State
		\Function {filterDetections}{\textit{detections}, \textit{targets}}
		\State remove \textit{detections} whose IOU with \textit{targets} in \texttt{tracked} state exceeds 0.5
		\EndFunction
		\State
		\Function {resolveTargetConflicts}{\textit{targets}}
		\State find all pairs of \texttt{tracked} \textit{targets} whose IOU $ > $ 0.7
		\State for each pair, remove the one that has shorter \texttt{tracked} streak
		\State if both have identical streak, remove the one having smaller IOU with maximally-overlapping detection			
		\EndFunction
		
	\end{algorithmic}
\end{algorithm}

\section{Deep MDP}
\label{deep_mdp}

\begin{figure}[!htbp]
	\includegraphics[width=0.9\textwidth]{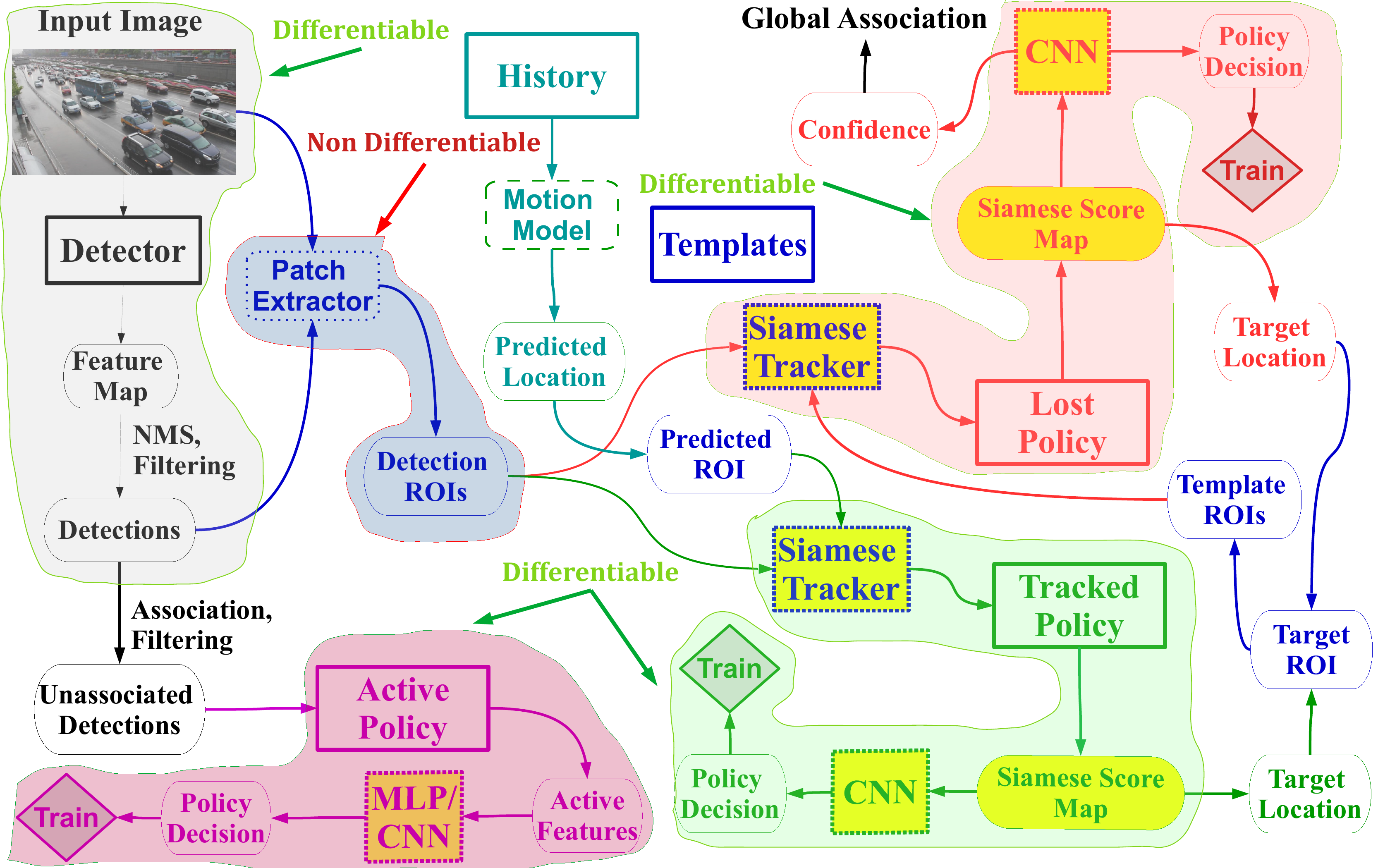}
	\includegraphics[width=0.9\textwidth]{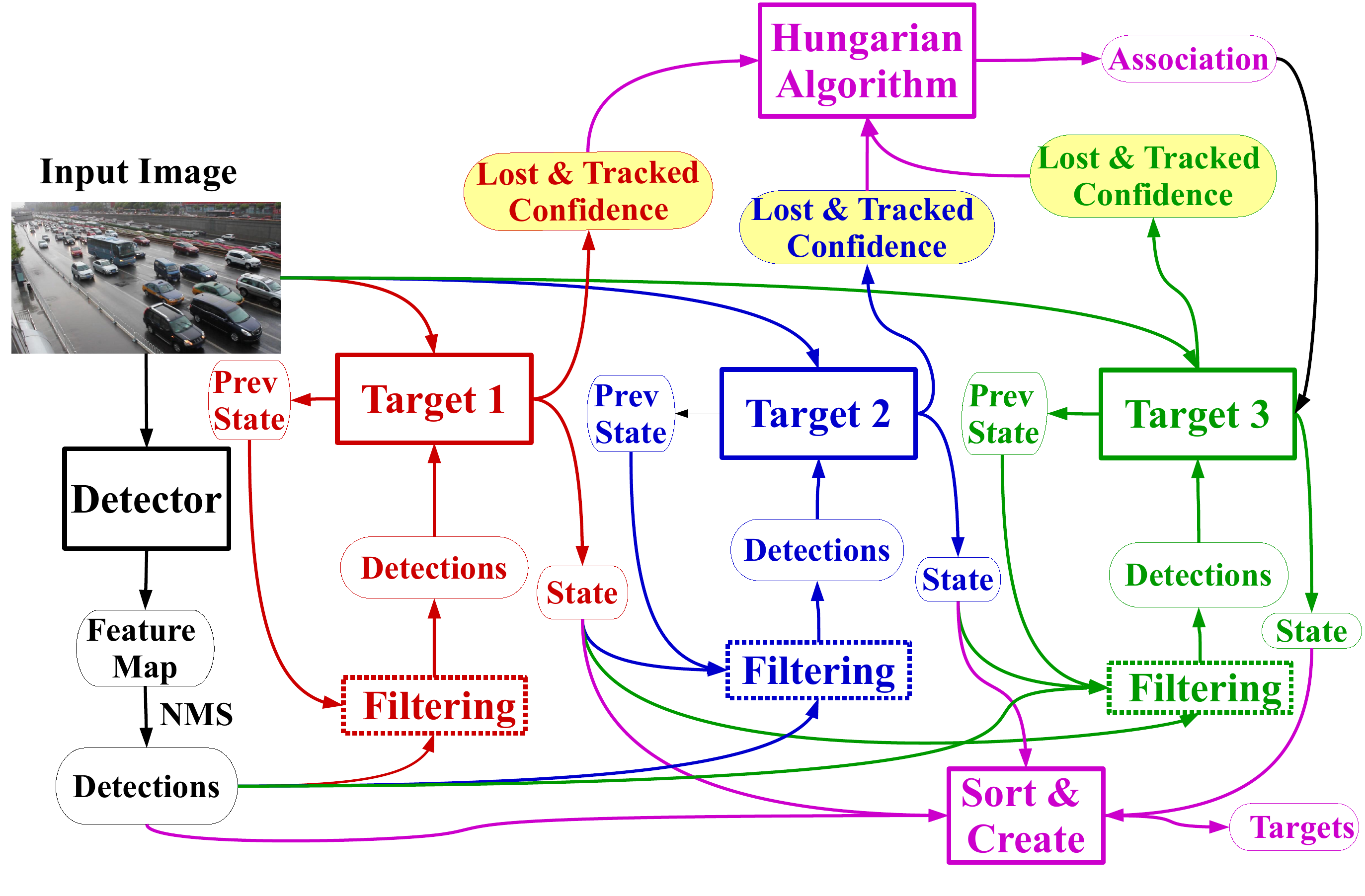}
	\caption{
		\small{
			Deep MDP Stage 1 block diagrams for (top) target-level and (bottom) global processing.
			Changed components with respect to MDP are shown with yellow-filled shapes.
			(Top) differentiable and non-differentiable parts of the pipeline are highlighted with green and red bordered regions filled with the colours of the respective module.
		}			
	}
	\label{fig-deep_mdp_1}
\end{figure}

\begin{figure}[!htbp]
	\includegraphics[width=\textwidth]{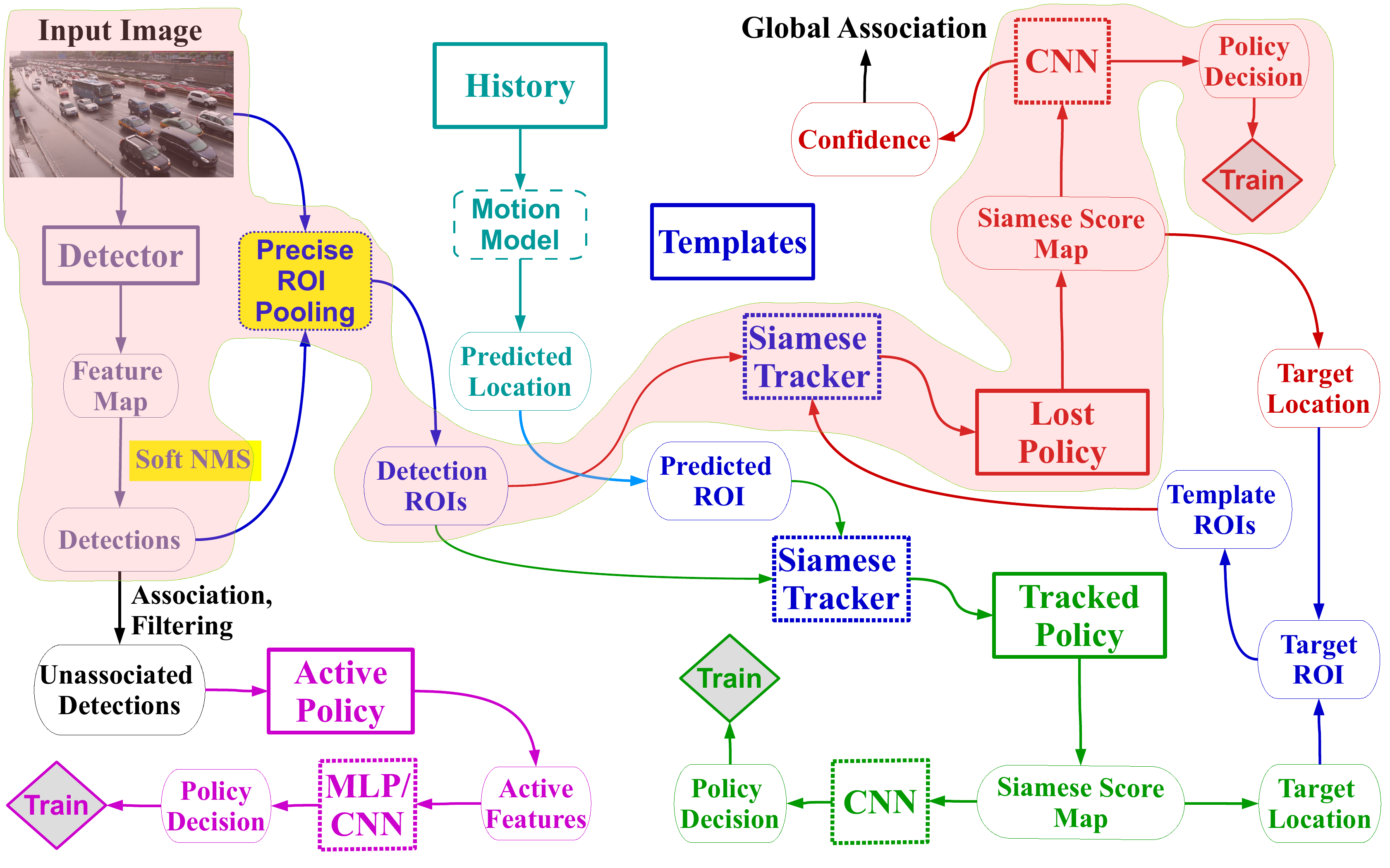}
	\includegraphics[width=\textwidth]{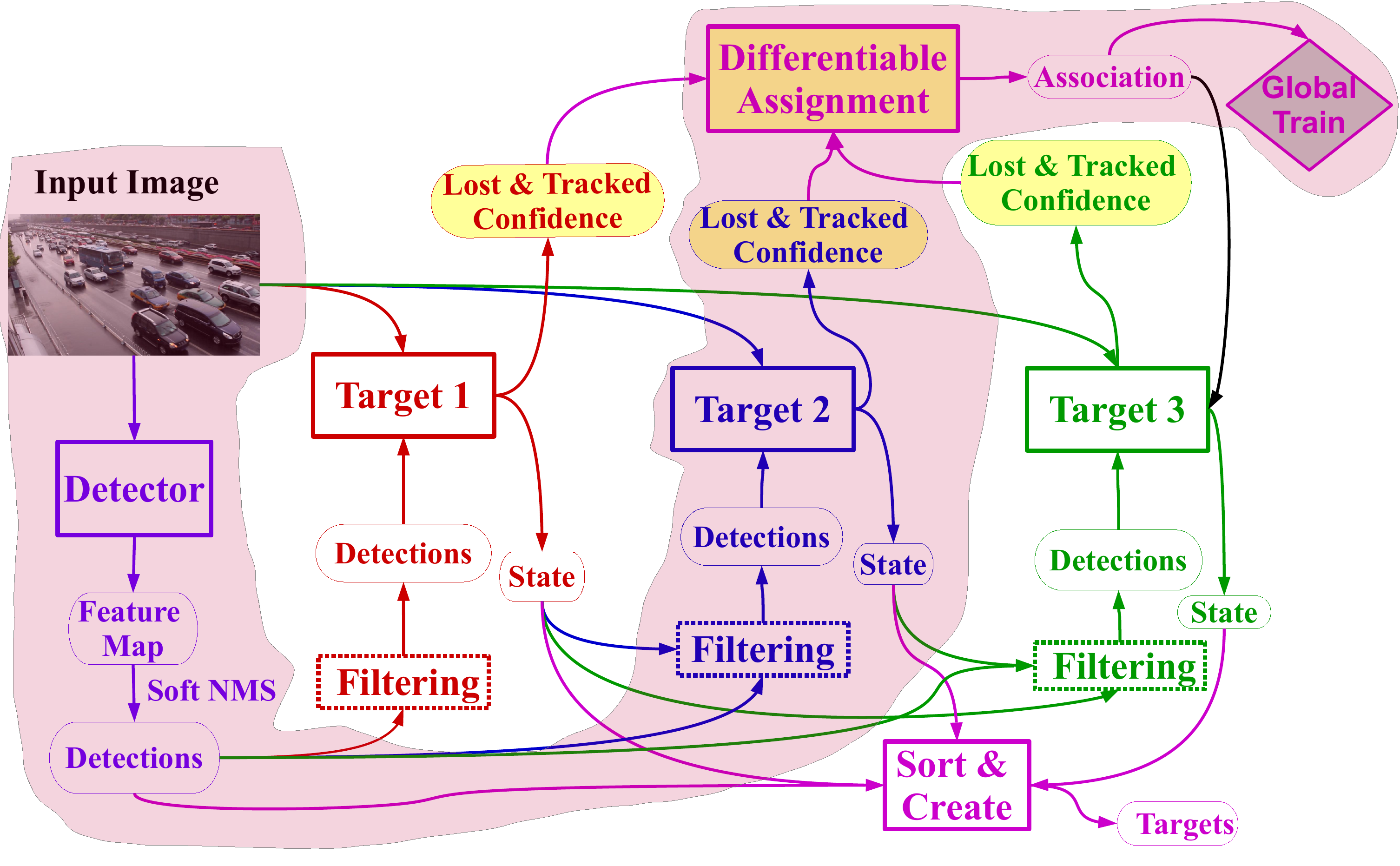}
	\caption{
		\small{
			Deep MDP block diagrams for (top) stage 2 target-level and (bottom) stage 3 global processing.
			Differentiable parts of the pipeline are highlighted with green bordered regions filled with the colour of \textit{lost} policy.
			(Top) a differentiable pipeline going through the \textit{tracked} policy can also be created but not shown for clarity
			(Bottom) differentiable pipelines going through targets 1 and 3 are not shown for clarity			
		}			
	}
	\label{fig-deep_mdp_2_3}
\end{figure}

I started
%on the task of
adapting MDP for deep learning with the twin objectives of improving its performance and creating an end-to-end differentiable MOT pipeline.
I devised a three stage plan to successively replace non-differentiable components
of the framework
%along the path from raw detections to the global association
by differentiable alternatives.
In Stage 1 (Fig. \ref{fig-deep_mdp_1}), I planned to replace the LK tracker with
end-to-end differentiable
%Siamese or correlation-based trackers,
deep learning based
trackers \cite{Zadeh21_sot_review,Jiao2021_sot_mot_review,Zhang2021_sot_review},
heuristics-based features with raw feature maps and SVM with an MLP or CNN classifier.
This would be followed in stage 2  (Fig. \ref{fig-deep_mdp_2_3} (top)) by replacing patch extractor with precise ROI pooling \cite{Jiang18_prroi_pool} to make the ROIs differentiable with respect to detection boxes, thus rendering the target-level pipeline fully differentiable.
Also, incorporating soft-NMS \cite{Bodla2017_soft_nms} would allow signals from all detections to be backpropagated.
Finally, I planned in stage 3 (Fig. \ref{fig-deep_mdp_2_3} (bottom)) to replace Hungarian association with a differentiable approximation (Sec. \ref{review-approx_diff}), thus making the global pipeline end-to-end trainable.
I have completed stage 1 implementation but have not succeeded in improving the performance.
%and have given up on this plan as impracticable after having recognized fundamental limitations in the MDP framework while diagnosing the poor performance.

The following sections
provide brief descriptions for
most of the ideas I have tried.
%towards this end.
%none of which
%that did not work
%and it should be clear, even if not explicitly stated, that none of these methods
%managed to significantly improve performance over the original MDP.
Quantitative results from some of these are included too, though it is sufficient to know that none managed to significantly improve performance over the original MDP.
Sec. \ref{deep_mdp-diagnostics} also contains details of some of the measures I took
%towards the end
to diagnose this poor performance which confirmed that the framework is simply not capable of significant
%performance
improvement without a complete redesign.

\subsection{Iterative Batch Training (IBT)}
\label{deep_mdp-ibt}

\begin{algorithm}[h]
	%\begin{algorithm}[t]
	\caption{
		Iterative Batch Training
	}				
	\label{alg-ibt}
	%	\scriptsize	
	\footnotesize
	\begin{algorithmic}[1]			
		\State \textit{max\_iters}, \textit{data\_from\_tester} $\gets$ parameters
		\State \textit{accumulative\_data}, \textit{train\_from\_scratch} $\gets$ parameters
		\State \textit{states} $\gets$ [active, tracked, lost]			
		\State \textit{training\_data} $\gets$ $ \emptyset $
		\For{\textit{iter} $\gets$ 1, \textit{max\_iters}}	
		\If{\textit{iter} = 1}	
		%				\textit{model}[\textit{state}] $\gets$ \texttt{relative oracle} $ \forall $ \textit{state} $ \in $ \textit{states} 
		\State set \textit{model} to \texttt{relative oracle} for all \textit{states}
		\Else
		\State
		%				\textit{model}[\textit{state}] $\gets$ \texttt{neural network} $ \forall $ \textit{state} $ \in $ \textit{states}
		set \textit{model} to \texttt{neural network} for all \textit{states}
		\If{\textit{train\_from\_scratch}}	
		\State initialize model weights randomly (or load pretrained weights) 			
		\Else
		\State load model weights from previous iteration
		\EndIf
		\EndIf
		\If{\textit{data\_from\_tester}}	
		\State \textit{new\_training\_data} $\gets$ DataFromTester(\textit{states})
		\EndIf
		\ForEach{\textit{state} $ \in $ \textit{states}}
		\If{\textbf{not} \textit{data\_from\_tester}}	
		\State load trained models for previous \textit{states} from current iteration 
		\State \textit{new\_training\_data} $\gets$ DataFromTrainer(\textit{state})
		\EndIf
		\If{ \textit{accumulative\_data}}	
		\State \textit{training\_data} $\gets$ \textit{training\_data} $ \cup $ \textit{new\_training\_data} 
		\Else
		\State \textit{training\_data} $\gets$ \textit{new\_training\_data} 
		\EndIf
		\State BatchTrain(\textit{training\_data}, \textit{state})
		\EndFor
		\State load trained models for all \textit{states} from current iteration			
		\State run Eval() to get classification accuracy on the test set
		\State run Test() to get tracking accuracy on the test set				
		\EndFor	
		
		\State
		\Function {DataFromTester}{\textit{states}}
		\State enable saving of samples from all policy decisions in all \textit{states}
		\State run \texttt{Tester} on training set
		\EndFunction	
		
		\State
		\Function {DataFromTrainer}{\textit{state}}
		\State enable saving of samples from all policy decisions in \textit{state}
		%		\State load trained model for \textit{state} from previous iteration
		%		\State load trained models for previous states from the same iteration
		\State set \textit{max\_passes} and \textit{max\_trials} to 1 so \texttt{Trainer} does a single pass over trajectories
		\State run \texttt{Trainer} on training set
		\EndFunction
		
		\State
		\Function {BatchTrain}{\textit{data}, \textit{state}}
		\State run batch training to train \textit{state} policy model on \textit{data}			
		\EndFunction
		
		\State
		\Function {Eval}{}	
		\State enable policy decision evaluation using GT 
		%		\State load trained models from current iteration for all states
		\State run \texttt{Trainer} on test set	
		\EndFunction	
		
		\State
		\Function {Test}{}	
		%		\State load trained models from current iteration for all states
		\State run \texttt{Tester} on test set
		\EndFunction		
	\end{algorithmic}
\end{algorithm}

After initial tests confirmed that incremental training would not work for deep networks,
%I first tried an asynchronous training process where 
I created this new training regime (Alg. \ref{alg-ibt}) to allow for batch training while still retaining several aspects of the original iterative training process (Alg. \ref{alg-mdp_training}).
As its name suggests, IBT involves batch-training of successively better models in an iterative process.
Each IBT iteration comprises the following four phases:
\begin{enumerate}[left=0pt,noitemsep,topsep=0pt]
	\item Data Generation: Run \texttt{Trainer} on the training set to get samples for each state:
	\begin{itemize}[left=0pt,noitemsep,topsep=0pt]
		\item First iteration: use relative oracle (Sec. \ref{deep_mdp-dummy}) and save all samples
		\item Second iteration onwards: use trained models from the previous iteration and save only incorrectly classified samples
	\end{itemize}
	\item Batch Training: Train each state on the samples collected in the data generation phase of the current iteration either by themselves or in addition to the samples collected in previous iterations.
	\item Evaluation: Run \texttt{Trainer} on the test set to evaluate classification accuracy
	\item Testing: Run \texttt{Tester} on the test set to evaluate tracking performance
\end{enumerate}
%where the first two
Note that
the first two phases
%data generation and batch training
are run individually for each state while
%evaluation and testing
the remaining two
are run only once per iteration
%jointly
%using the latest trained models
%for all states
with the latest trained models
loaded for all states.
\subsubsection*{Data from Tester}
I also collected training samples directly
by running
%from
the \texttt{Tester} to mitigate the aforementioned disconnect between training and inference caused by
%using data generated
generating data
in the \texttt{Trainer}
to train models that are deployed in the \texttt{Tester}.
Contrary to standard data generation,
this too is run only once per iteration to generate samples jointly for all states.

\subsection{Tracking}
\label{deep_mdp-tracking}
\begin{table*}[!htbp]
	\centering
	\caption{
		\small{
			Comparing
			%		the
			DETRAC tracking
			performance
			%		of
			with
			different patch trackers.
			Siamese-FC, DA-SiamRPN and DiMP were combined with a ResNet18 CNN trained on raw score maps with IBT.
			LK was combined with SVM and trained using the original \texttt{Trainer}.
			These results give a good overall idea of the relative performance between MDP and deep MDP.
			% Results were generated on sequences 30-50 of DETRAC.
		}
	}
	\begin{tabular}{|c|ccc|cc|}
		\hline
		\multirow{2}{*}{\textbf{Tracker}} & \multicolumn{3}{c|}{\textbf{higher is better}}                                                      & \multicolumn{2}{c|}{\textbf{lower is better}}       \\ \cline{2-6} 
		& \multicolumn{1}{c|}{\textbf{MOTA (\%)}} & \multicolumn{1}{c|}{\textbf{MOTP (\%)}} & \textbf{MT(\%)} & \multicolumn{1}{c|}{\textbf{ML(\%)}} & \textbf{IDS} \\ \hline
		\textbf{Siamese-FC}               & \multicolumn{1}{c|}{83.508}            & \multicolumn{1}{c|}{55.219}            & 79.963       & \multicolumn{1}{c|}{3.035}        & 2902         \\ \hline
		\textbf{DA-SiamRPN}               & \multicolumn{1}{c|}{78.390}            & \multicolumn{1}{c|}{41.389}            & 65.469       & \multicolumn{1}{c|}{5.451}        & 2610         \\ \hline
		\textbf{DiMP}                     & \multicolumn{1}{c|}{84.071}            & \multicolumn{1}{c|}{56.006}            & 77.145       & \multicolumn{1}{c|}{3.654}        & 2661         \\ \hline
		\textbf{LK}                       & \multicolumn{1}{c|}{83.708}            & \multicolumn{1}{c|}{51.937}            & 76.525        & \multicolumn{1}{c|}{3.561}        & 2233         \\ \hline
	\end{tabular}
	\label{tab-trackers}
\end{table*}
The first tracker I added was
the popular
Siamese-FC \cite{Bertinetto2016_siamese_fc} which failed to outperform LK (Table \ref{tab-trackers}) and was also too slow for extensive testing.
I followed it with the newer and much faster DA-SiamRPN \cite{Zhu2018_da_siam_rpn} which I used for a majority of the subsequent testing due to its high speed and relatively low memory requirements.
However, this too failed to improve over LK and was in fact slightly worse than Siamese-FC in most cases.
I therefore also experimented with several improved Siamese trackers including SiamRPN \cite{Li2018_siam_rpn}, SiamRPN++ \cite{Li19_SiamRPNpp}, SiamVGG \cite{Li2019_Siam_VGG} and SiamDW \cite{Zhang2019_siam_dw} by integrating the SiameseX framework \cite{siamx_github} with deep MDP.
%None of these provided any performance improvements either so I also tried
Finally, I tried
several correlation-based trackers including ECO \cite{Danelljan2017_eco}, ATOM \cite{Danelljan2019_ATOM}, DiMP \cite{Bhat2019_dimp} and PrDiMP \cite{Danelljan2020_prdimp} implemented as part of the pytracking \cite{pytracking_github} framework.
%These too failed to provide any performance improvements.
%The only thing I omitted is to train one of these trackers on the MOT dataset on which they were tested but I did test on a variety of vehicle and pedestrian tracking scenarios which are quite similar to the SOT benchmarks on which the pre-trained models that I used were trained.

\subsubsection{Continuous Tracking Mode (CTM)}
\label{deep_mdp-ctm}
MDP uses discontinuous ROI-to-ROI tracking (Sec.  \ref{mdp-lk}) that is not suitable for the above trackers since these are designed for continuous long-term tracking.
Therefore, I added CTM to the \texttt{tracked} state to
bypass the pairwise tracking between templates and predicted locations 
and instead
track an object continuously from the first frame in which it is added.
Tracking stops when the object moves from \textit{tracked} to \textit{lost} and the tracker is reinitialized when the object transitions back to \textit{tracked} to account for the frame discontinuity.
%I have not extended CTM to the \texttt{lost} state since tracking an occluded object would not make sense.
CTM does improve performance
%over pairwise tracking
somewhat but not enough to outperform the baseline.

\subsection{Classification}
\label{deep_mdp-classification}
I first tried using relatively small custom-designed MLPs with 3 to 7 layers for both \textit{active} and \textit{lost} policies followed by CNNs with 3 to 8 layers for \textit{lost} and \textit{tracked}.
%These small networks worked fairly well for \textit{active} policy but not for \textit{lost}.
These did outperform SVM using handcrafted features from both the original MDP as well as those extracted from the Siamese score maps (sec. \ref{deep_mdp-feature})
%that I designed myself
but did not lead to any improvement in the overall MOT performance itself.
%These worked quite well for \textit{active} policy with the existing handcrafted features.
I then experimented with several off-the shelf networks including VGG-13 \cite{Simonyan15_vgg}, MobileNetv2 \cite{Sandler2018_MobileNetV2}, Inceptionv3 \cite{Szegedy2016_Inceptionv3}, ResNet18/50/101 \cite{He16_resnet} and ResNext101 \cite{Xie17_resnext}, though mostly for \textit{lost} and \textit{tracked} policies since \textit{active} policy does not appear to have as much impact on performance.
I also
%tried a modified Siamese-style \cite{Koch2015_Siamese} version of ResNet 
modified ResNet to
%that can
take two image patches as input and process them in parallel through 
%that uses shared layers to process two input images in parallel
Siamese-style \cite{Koch2015_Siamese} shared layers
%in order
%to directly feed the two image patches being compared as input.
so that the two ROIs being compared can be fed directly as input
without having to go through the tracker.

\subsubsection*{Model Sharing}
Detailed examination of \textit{lost} and \textit{tracked} training samples and decision scenarios suggested
%to me
that both policies require
%more or less the same sort of
similar
discriminative abilities since both need to decide if two patches correspond to the same object.
Therefore, I implemented a model sharing functionality where a model trained for one of the policies can also be used for the other.

\begin{table*}[t]
	\centering
	\caption{
		Quantitative details of datasets used for evaluation.
		%		Test set annotations are unavailable for MOT 2015 / 2017, CTMC and CTC though trajectory stats are available for the first two.
	}
	\begin{tabular}{|cc|c|c|c|c|}
		\hline
		\multicolumn{2}{|c|}{\textbf{Dataset}}                                                            & \textbf{Subset} & \textbf{Sequences} & \textbf{Frames   (K)} & \textbf{Trajectories} \\ \hline
		\multicolumn{1}{|c|}{\multirow{6}{*}{\textbf{Pedestrian}}} & \multirow{3}{*}{\textbf{MOT   2015}} & \textbf{train}  & 10                 & 5.5                   & 570                   \\ \cline{3-6} 
		\multicolumn{1}{|c|}{}                                     &                                      & \textbf{test}   & 10                 & 5.8                   & 721                   \\ \cline{3-6} 
		\multicolumn{1}{|c|}{}                                     &                                      & \textbf{total}  & 20                 & 11.3                  & 1291                  \\ \cline{2-6} 
		\multicolumn{1}{|c|}{}                                     & \multirow{3}{*}{\textbf{MOT 2017}}   & \textbf{train}  & 7                  & 5.3                   & 796                   \\ \cline{3-6} 
		\multicolumn{1}{|c|}{}                                     &                                      & \textbf{test}   & 7                  & 5.9                   & 2354                  \\ \cline{3-6} 
		\multicolumn{1}{|c|}{}                                     &                                      & \textbf{total}  & 14                 & 11.2                  & 3150                  \\ \hline
		\multicolumn{1}{|c|}{\multirow{5}{*}{\textbf{Vehicle}}}    & \textbf{GRAM}                        & \textbf{total}  & 3                  & 40.3                  & 751                   \\ \cline{2-6} 
		\multicolumn{1}{|c|}{}                                     & \textbf{IDOT}                        & \textbf{total}  & 13                 & 111.7                 & 1460                  \\ \cline{2-6} 
		\multicolumn{1}{|c|}{}                                     & \multirow{3}{*}{\textbf{DETRAC}}     & \textbf{train}  & 60                 & 83.8                  & 5937                  \\ \cline{3-6} 
		\multicolumn{1}{|c|}{}                                     &                                      & \textbf{test}   & 40                 & 56.3                  & 2319                  \\ \cline{3-6} 
		\multicolumn{1}{|c|}{}                                     &                                      & \textbf{total}  & 100                & 140.1                 & 8256                  \\ \hline
		\multicolumn{1}{|c|}{\multirow{6}{*}{\textbf{Cell}}}       & \multirow{3}{*}{\textbf{CTMC}}       & \textbf{train}  & 47                 & 80.3                  & 1615                  \\ \cline{3-6} 
		\multicolumn{1}{|c|}{}                                     &                                      & \textbf{test}   & 39                 & 72                    & N/A                   \\ \cline{3-6} 
		\multicolumn{1}{|c|}{}                                     &                                      & \textbf{total}  & 86                 & 152.3                 & N/A                   \\ \cline{2-6} 
		\multicolumn{1}{|c|}{}                                     & \multirow{3}{*}{\textbf{CTC}}        & \textbf{train}  & 20                 & 8                     & 4299                  \\ \cline{3-6} 
		\multicolumn{1}{|c|}{}                                     &                                      & \textbf{test}   & 20                 & 8.1                   & N/A                   \\ \cline{3-6} 
		\multicolumn{1}{|c|}{}                                     &                                      & \textbf{total}  & 40                 & 16.1                  & N/A                   \\ \hline
	\end{tabular}
	\label{tab-datasets}
\end{table*}

\subsection{Feature Extraction}
\label{deep_mdp-feature}
%I initially tried using the original MDP handcrafted features with my custom-designed MLPs which worked well enough compared to SVM but showed signs of overfitting which I interpreted as being due to limited representational power of these features.
Since passing the raw Siamese score maps directly to the classifier works only with CNNs,
I also devised a few handcrafted features from these maps for use with MLPs.
These are based on the heuristic that successful tracking produces score maps that have a single and relatively concentrated or sharply defined region of maximum.
This tends to become more diffused or gets supplemented with additional
%distinct
regions of maxima when the tracking is unsuccessful.
The simplest such feature is generated by concatenating the maximum of each row and column in the score map.
Another similar way is to concatenate a certain number of the largest values in the score map with optional non-maximum suppression (NMS) to suppress high values in the neighbourhood of each local maximum. 
%This is designed on the intuition that successful tracking would produce a score map where there is a sharp drop in scores after the single or top few highest values while unsuccessful tracking would have relatively gradual drop-off.
A more sophisticated variant involves computing statistics (e.g. mean, median, minmum and maximum) from concetric neighbourhoods of increasing radii around the point of maximum of the score map.

I next used the raw score maps directly as input to the various CNNs I experimented with.
Finally, I tried feeding the two image patches themselves as input to Siamese-style ResNet CNNs, thus bypassing the tracker in feature extraction.
For feature summarization over templates, I
%experimented with
tried
%both the original anchor-template selection heuristics as well as
taking the
elementwise mean of features from all templates
as well as
%and use its features.
%I also tried
concatenating all features
%from all templates
into a single multichannel feature fed directly into the classifier
in order to
%so as to
learn summarization in the network too.
%so that summarization can be learnt too.
%the latter can learn to perform summarization too.
%\subsubsection{Training}

%\subsubsection*{Handcrafted Siamese Features}
%\subsubsection*{Raw Features / multichannel / siamese classifier}
%

\subsection{Class Imbalance}
\label{deep_mdp-imbalance}
\begin{table*}[!htbp]
	\centering
	\caption{
		\small{
			Number of positive and negative training samples collected
			using IBT data from tester module 
			along with corresponding majority to minority class ratios.
		}
	}
	\begin{tabular}{|cccc|cccc|}
		\hline
		\multicolumn{4}{|c|}{\textbf{GRAM}}                                                                                                       & \multicolumn{4}{c|}{\textbf{DETRAC}}                                                                                                    \\ \hline
		\multicolumn{1}{|c|}{\textbf{policy}}  & \multicolumn{1}{c|}{\textbf{negative}} & \multicolumn{1}{c|}{\textbf{positive}} & \textbf{ratio} & \multicolumn{1}{c|}{\textbf{policy}} & \multicolumn{1}{c|}{\textbf{negative}} & \multicolumn{1}{c|}{\textbf{positive}} & \textbf{ratio} \\ \hline
		\multicolumn{1}{|c|}{\textbf{active}}  & \multicolumn{1}{c|}{117583}            & \multicolumn{1}{c|}{425}               & 276.7          & \multicolumn{1}{c|}{active}          & \multicolumn{1}{c|}{28505}             & \multicolumn{1}{c|}{287}               & 99.3           \\ \hline
		\multicolumn{1}{|c|}{\textbf{tracked}} & \multicolumn{1}{c|}{780}               & \multicolumn{1}{c|}{44077}             & 56.5           & \multicolumn{1}{c|}{tracked}         & \multicolumn{1}{c|}{627}               & \multicolumn{1}{c|}{45842}             & 73.1           \\ \hline
		\multicolumn{1}{|c|}{\textbf{lost}}    & \multicolumn{1}{c|}{2553}              & \multicolumn{1}{c|}{549}               & 4.7            & \multicolumn{1}{c|}{lost}            & \multicolumn{1}{c|}{1865}              & \multicolumn{1}{c|}{676}               & 2.8            \\ \hline
	\end{tabular}
	\label{tab-class_imbalance}
\end{table*}

Significant class imbalance is a major issue in deep MDP training (Table \ref{tab-class_imbalance}) that prevents the models from converging to good optima and instead leads to overfitting to a particular class.
%For example, the most common type of failure I found was in overfitted \textit{lost} policies that classified nearly all samples as negative, thus rapidly leading to an overflow of \textit{lost} targets.
Following are the main strategies I tested to address
%the class imbalance
this
issue:
\begin{itemize}[left=0pt,noitemsep,topsep=0pt]
	\item Focal loss \cite{Lin2017_retinanet}: Dynamically scale the cross-entropy loss to reduce the weight given to high confidence samples so that training is not overwhelmed by easy to classify examples.
	\item  Online Hard Example Mining (OHEM) \cite{Shrivastava2016_ohem}: Select a certain ratio of samples with the highest loss and compute the loss using only these so as to concentrate training on the examples that are the most difficult for the network to classify.
	\item Under sampling: Select a different random subset of the majority class of the same size as the minority class in each epoch
	\item Over sampling: Randomly duplicate samples from the minority class so that it has the same size as the majority class
	\item Probabilistic sampling \cite{imbalanced_sampler_github}: Draw samples probabilistically from the entire training set such that the probability of each sample being drawn is inversely proportional to its class frequency.
	\item Synthetic samples: Using each real box as an anchor, generate randomly shifted and resized boxes around it and choose synthetic samples from these by
	IOU-based
	thresholding:
	\begin{itemize}[left=0pt,noitemsep,topsep=0pt]
		\item Negative sample: $ 0.1 < $ IOU with anchor $ < 0.3 $, IOU with other real boxes $< 0.3$, IOU with other synthetic boxes $< 0.5$
		\item Positive sample: $ 0.5 < $ IOU with anchor $ < 0.8 $, IOU with other real boxes $< 0.8$, IOU with other synthetic boxes $< 0.5$
	\end{itemize}
\end{itemize}

\section{Diagnostics}
\label{deep_mdp-diagnostics}
This section briefly describes some of the many measures I took to diagnose the poor performance of deep MDP along with
%details of
%the
%corresponding
datasets and metrics
I used throughout.
%for the purpose as well as for general testing.

\subsection{Datasets}
\label{deep_mdp-datasets}
%The main datasets I have used are
%MOT 2015 \cite{Leal15_MOT15} and 2017 \cite{Milan16_MOT16}
%for pedestrian tracking,
%IDOT \cite{Jin2017_idot},  GRAM \cite{GuerreroGmezOlmedo2013_GRAM} and DETRAC \cite{Wen2020_DETRAC}
%for vehicle tracking,
%and
%CTMC \cite{Anjum2020_CTMC} and CTC \cite{Ulman2017_CTC} 
%for cell tracking.
Following are the main datasets I have used:
\begin{itemize}[noitemsep,topsep=0pt]
	\item Pedestrian: MOT 2015 \cite{Leal15_MOT15}, MOT 2017 \cite{Milan16_MOT16}
	\item Vehicle: DETRAC \cite{Wen2020_DETRAC}, IDOT \cite{Jin2017_idot}, GRAM \cite{GuerreroGmezOlmedo2013_GRAM}
	\item Cell: CTMC \cite{Anjum2020_CTMC}, CTC \cite{Ulman2017_CTC}
\end{itemize}
Table \ref{tab-datasets} provides quantitative details for these.

%\begin{table}[!htbp]
%	\centering
%	\caption{
	%		Detection quality for some of the datasets
	%	}
%	\begin{tabular}{|c|c|c|c|c|}
	%		\hline
	%		\textbf{dataset}                   & \textbf{detector} & \textbf{recall} & \textbf{precision} & \textbf{mAP} \\ \hline
	%		\textbf{MOT 2015}                  & ACF               & 52.16           & 60.32              & 39.46        \\ \hline
	%		\multirow{3}{*}{\textbf{MOT 2017}} & FRCNN             & 31.24           & 94.53              & 31.18        \\ \cline{2-5} 
	%		& SDP               & 39              & 96.43              & 38.89        \\ \cline{2-5} 
	%		& DPM               & 25.1            & 64.4               & 23.27        \\ \hline
	%		\textbf{GRAM}                      & RFCN + AE         & 66.7            & 27.21              & 46.55        \\ \hline
	%		\textbf{DETRAC}                    & FRCNN             & 80.7            & 66.13              & 68.43        \\ \hline
	%	\end{tabular}
%	\label{tab-detections}
%\end{table}
\subsection{Metrics}
For most of the testing, I used the CLEAR MOT metrics MOTA and MOTP \cite{Bernardin2008_mota} along with a few other popular metrics from the MOT Challenge \cite{Leal15_MOT15} including Mostly Tracked (MT), Mostly Lost (ML) and IDS.
The inadequacies and mutual inconsistencies of these metrics are well known \cite{Leichter2013_mota_criticism,Milan2013_mota_criticism,Bento2016_mota_criticism,Ristani2016_idf1,LealTaix17T_mot_review,Luiten2021_HOTA} and it was possible that these could be at least partially responsible for the failure of deep MDP to outperform MDP.
Therefore,
towards the end,
I also experimented with the
%then recently released
HOTA metrics \cite{Luiten2021_HOTA} that are supposedly less dependent on the detection quality.
However, I did not find any significant difference in relative performance of different models between HOTA and the earlier metrics.
%except perhaps MOTA which favored MDP over deep MDP slightly more than all others.
Following are brief definitions for these metrics:
\begin{enumerate}[topsep=0pt]
	\item MOTA = $ 1 - \frac{|\text{FN}| + |\text{FP}| + \text{IDS}}{|\text{GT}|} $ 
	\item MOTP: Mean dissimilarity between TPs and their corresponding GTs
	\item MT: Fraction of trajectories with $ > 80\% $ boxes tracked correctly
	\item ML: Fraction of trajectories with $ < 20\% $ boxes tracked correctly
	\item IDS: Number of times target ID changes in the middle of its trajectory
	\item HOTA: $ \sqrt{\text{DetA}\cdot\text{AssA}} $ where
	\begin{itemize}[left=0pt,topsep=0pt,label=\textendash]
		\item $ \text{DetA} = \text{Detection Accuracy} = \frac{|\text{TP}|}{|\text{TP}| + |\text{FP}| + |\text{FN}|} $
		\item $ \text{AssA} = \text{Association Accuracy} = \frac{1}{|\text{TP}|}\sum\limits_{c\in\{\text{TP}\}} \frac{|\text{TPA}|}{|\text{TPA}| + |\text{FPA}| + |\text{FNA}|} $
		\item TPA, FPA and FNA are similar to TP, FP and FN except measured at the level of trajectories instead of boxes.
	\end{itemize}
\end{enumerate}

\subsection{Validating Benefit of End-to-End Training}

\begin{figure*}[t]
	\centering
	\includegraphics[width=0.32\textwidth]{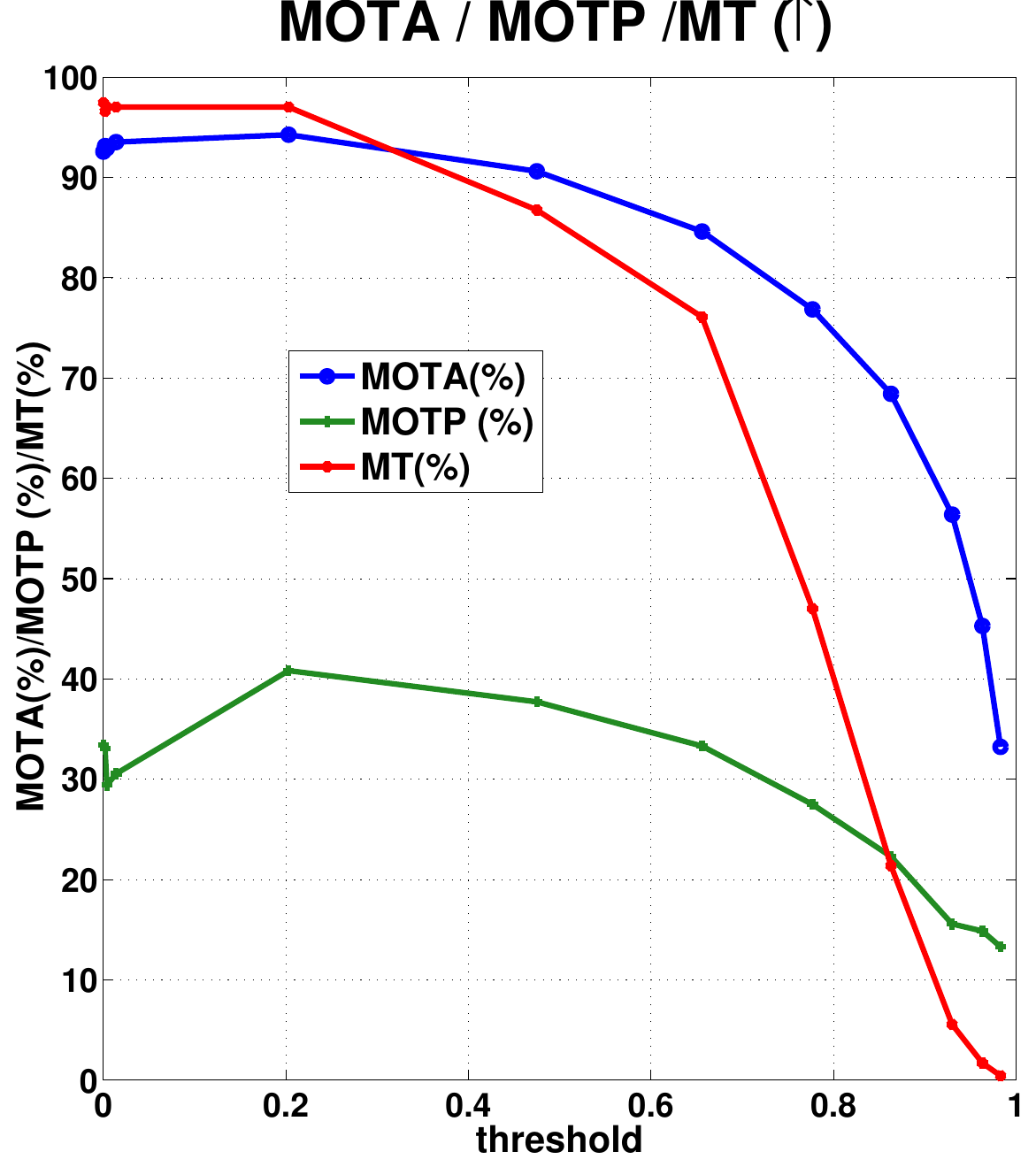}
	\includegraphics[width=0.32\textwidth]{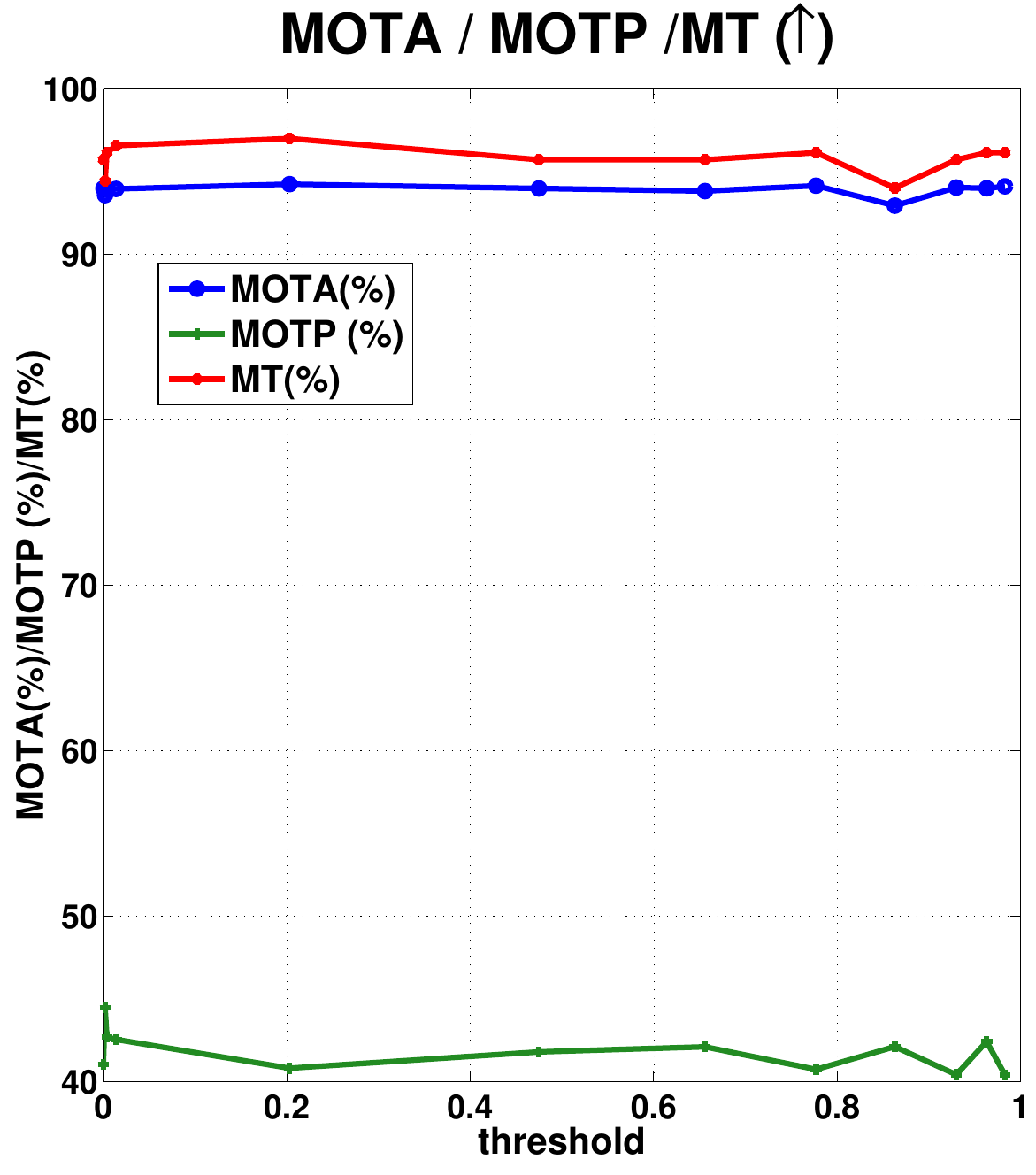}
	\includegraphics[width=0.32\textwidth]{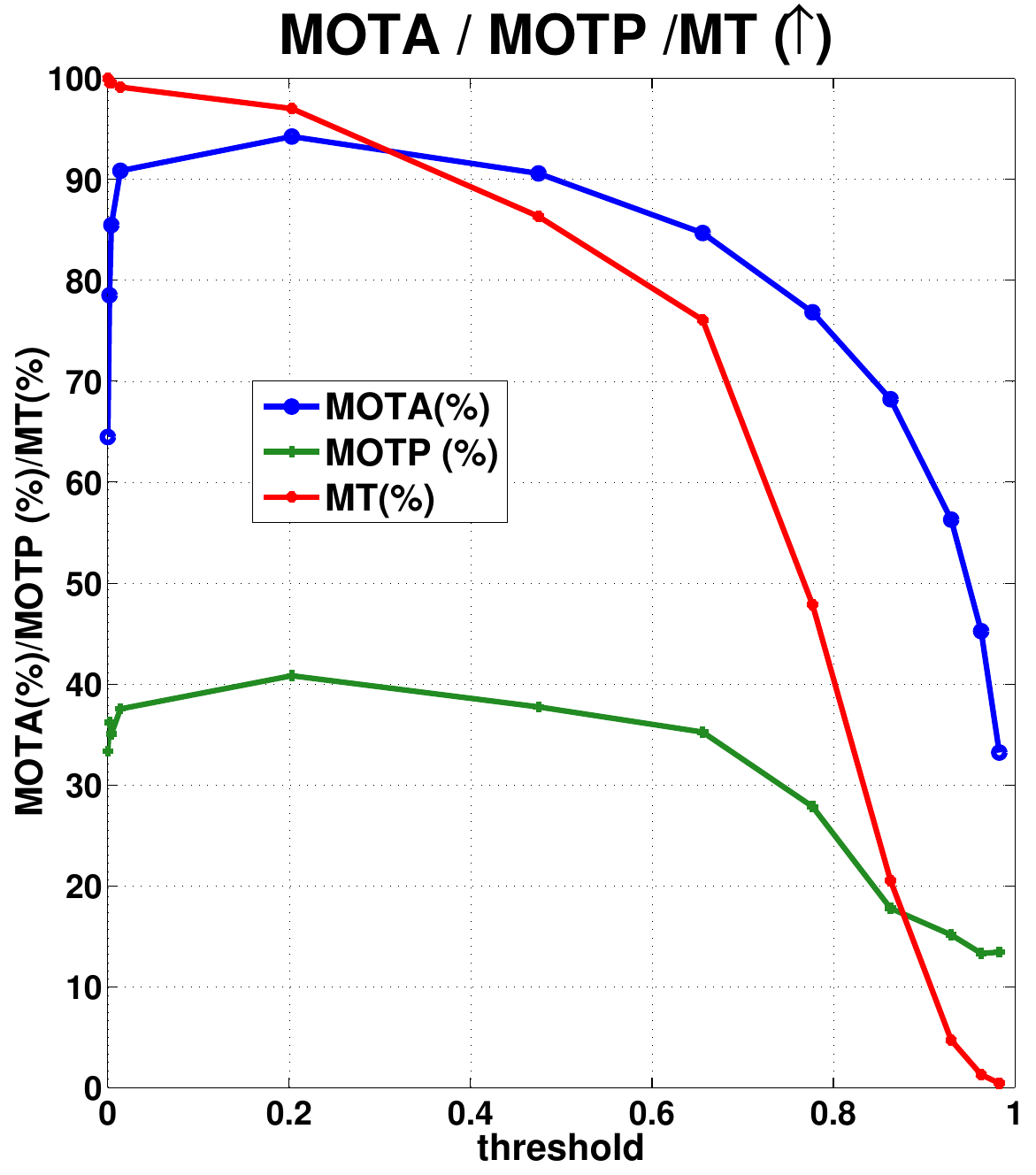}
	
	\includegraphics[width=0.32\textwidth]{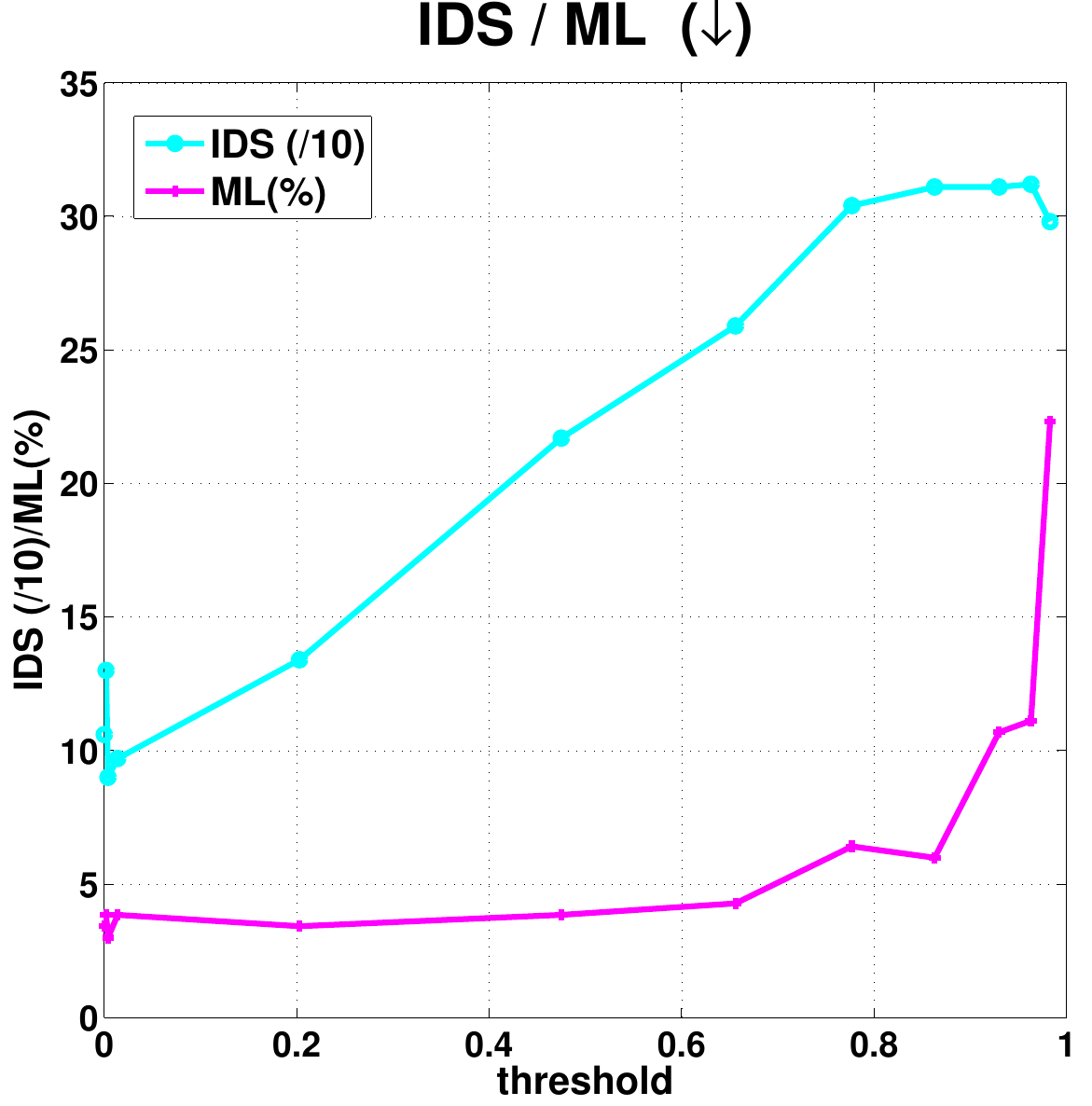}
	\includegraphics[width=0.32\textwidth]{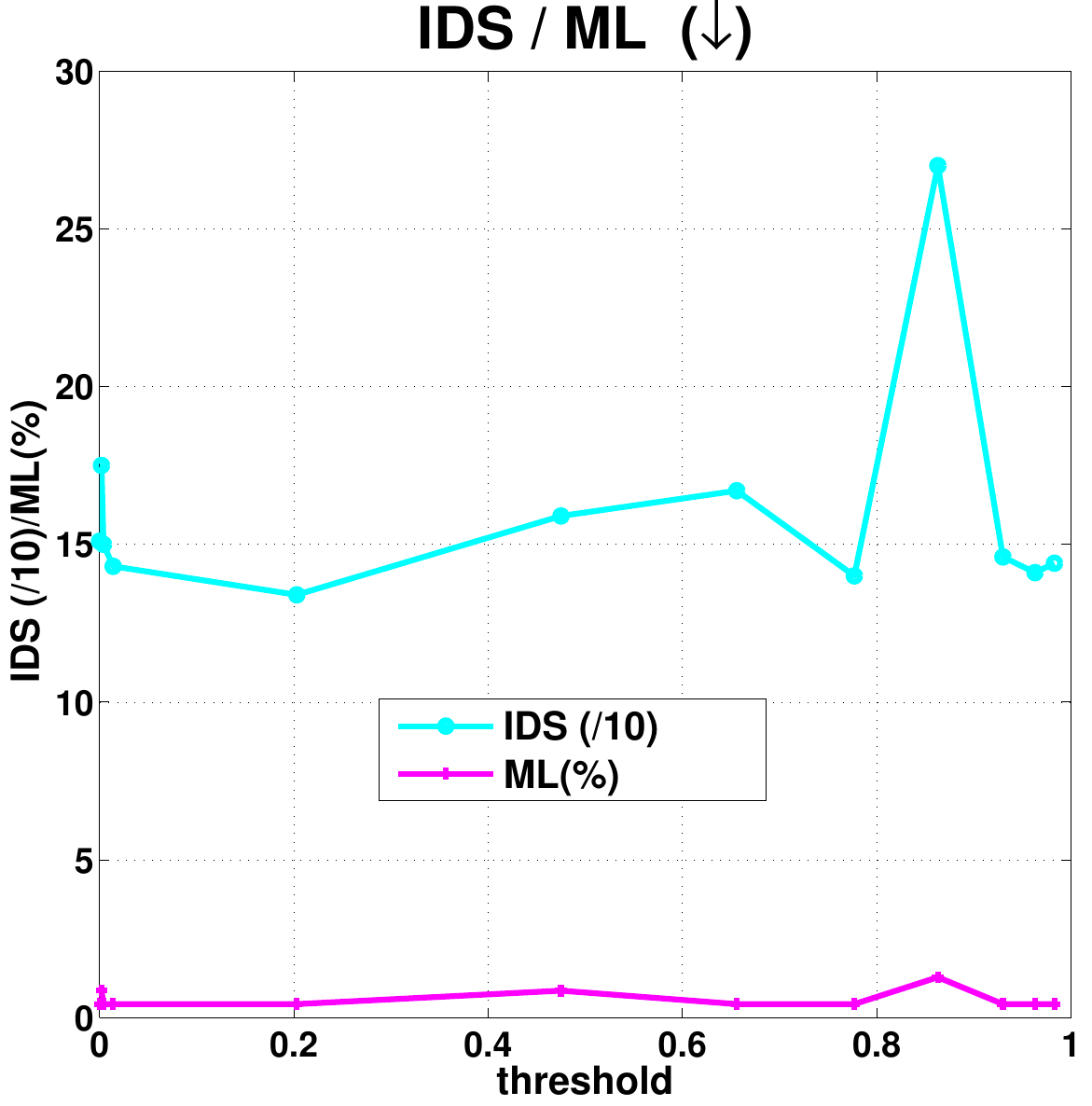}
	\includegraphics[width=0.32\textwidth]{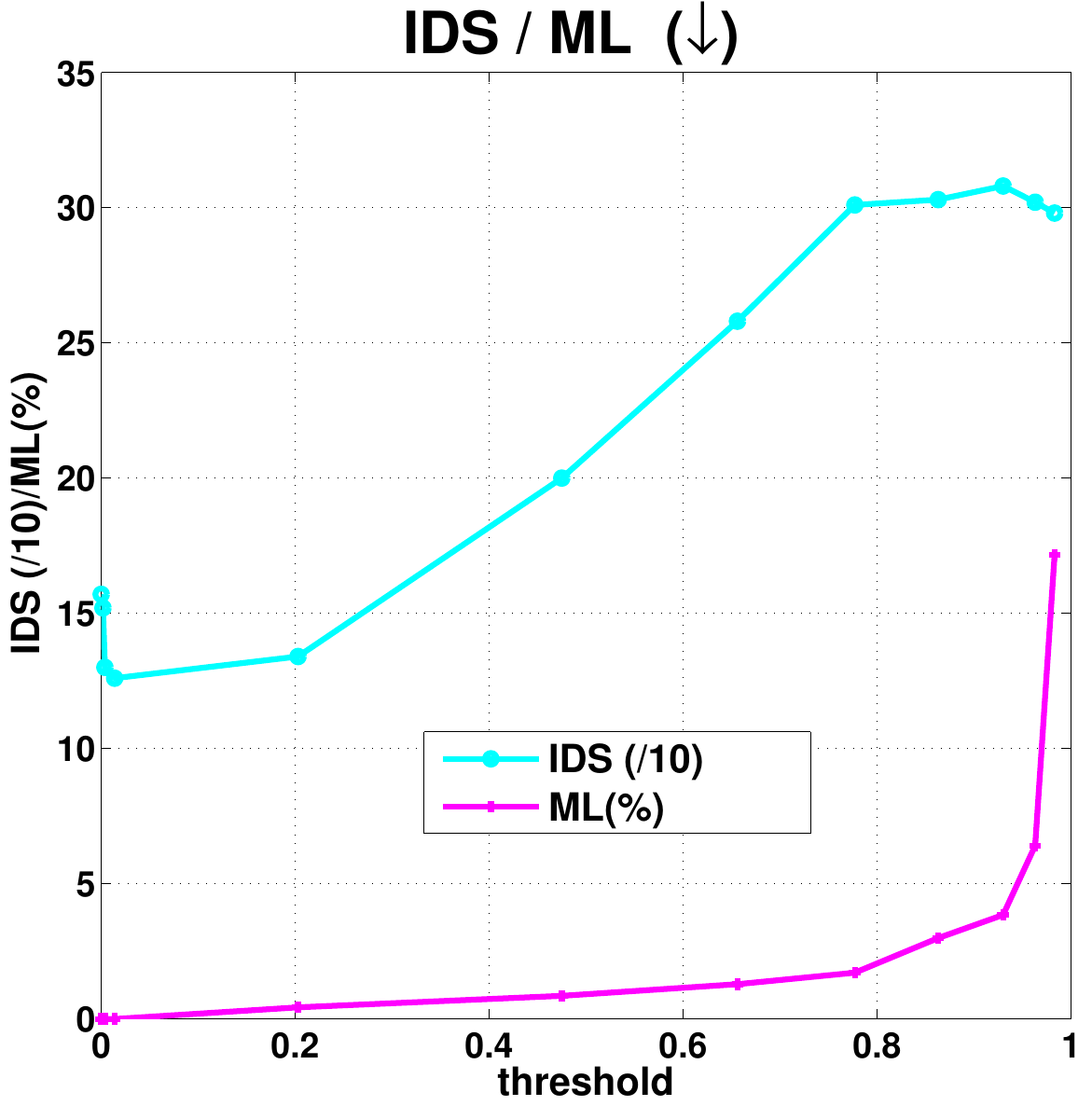}
	\caption{
		\small{
			Results of testing three combinations of trackers and detectors with varying recall-precision trade-offs: (from left to right) each tracker with its own detector, all trackers with the best detector and the best tracker with all the detectors.
			The x-axis in the left and right plots corresponds to detectors with different tradeoffs
			(obtained by varying the confidence thresholds as shown in Fig. \cite{fig-yolo_rec_prec})		
			%The x-axis
			while that
			in the center plot corresponds to different MDP models trained with these detectors.
			Top and bottom rows show metrics for which higher and lower is better respectively.
			IDS(/10) in the latter means that the actual IDS is 10 times the values plotted.\\
			These results were generated using standard MDP with LK and SVM but similar plots were obtained using DA-SiamRPN and DiMP with ResNet18 too so they can be taken to represent module-independent properties of the framework itself.
			It can be seen that the left and right plots are nearly identical, thus indicating that MDP training process hardly learns anything from the detector and is virtually incapable of adapting itself to the detector characteristics to any degree.
			The drop in tracking performance with increase in threshold also highlights the well-known fact that MOT performance is far more sensitive to FNs than FPs.
			Finally, the near-perfect performance of all trackers in the center plot shows the heavy dependence of MDP on detection quality.
		}
	}
	\label{fig-yolo_sweep_results}
\end{figure*}

\begin{figure}[!htbp]
	\centering
	\includegraphics[width=0.38\textwidth]{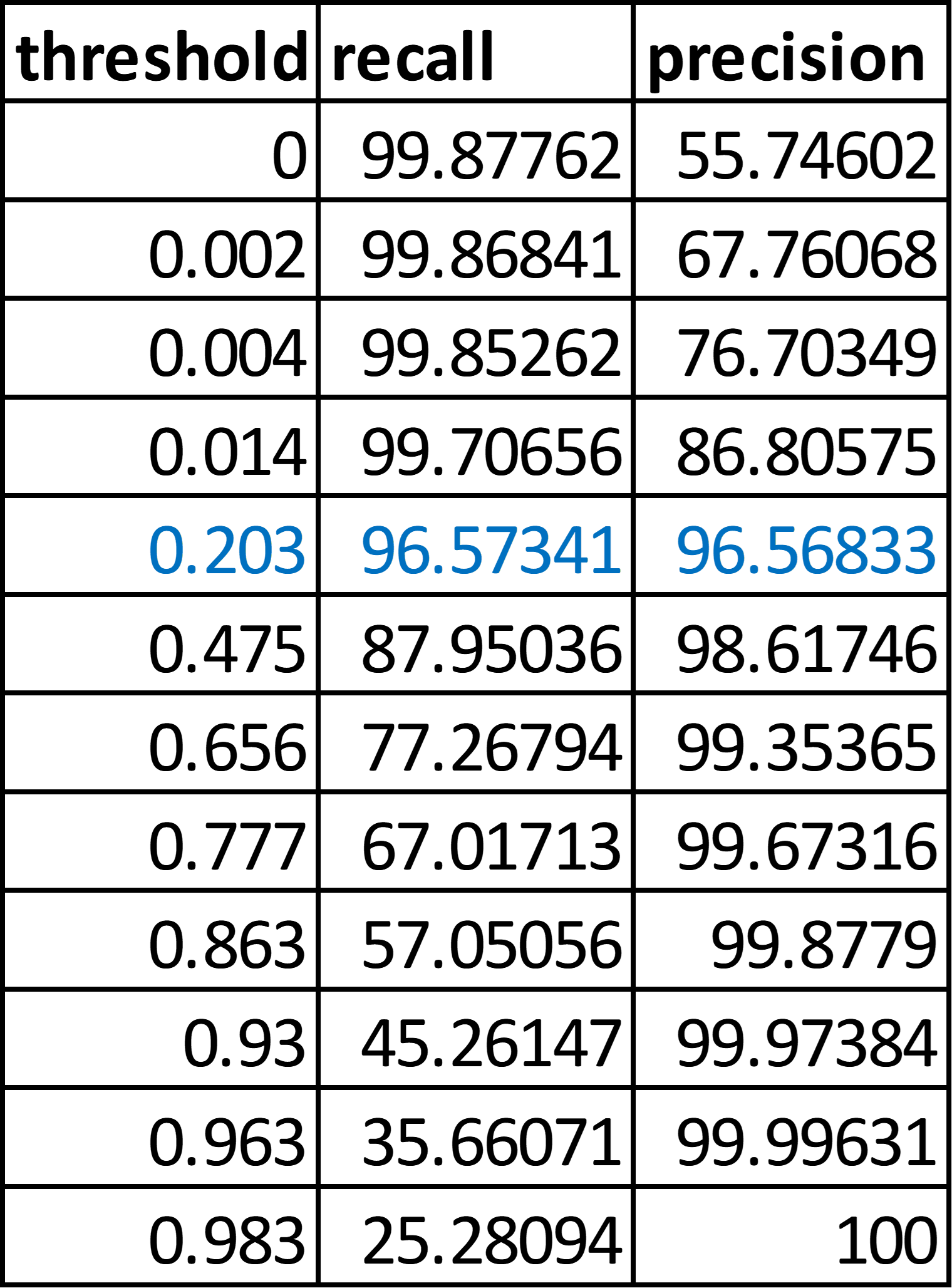}
	\includegraphics[width=0.48\textwidth]{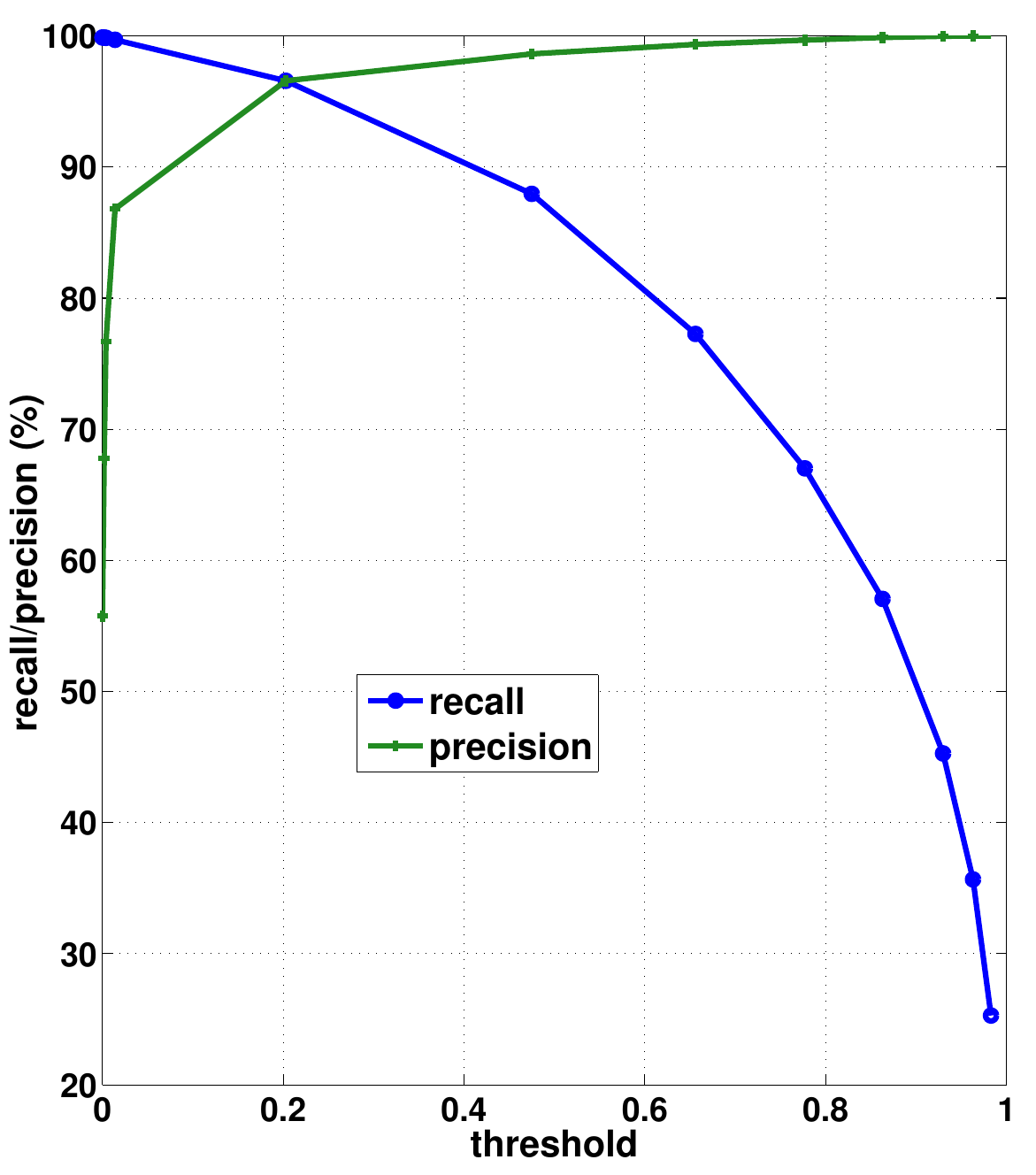}
	\caption{
		\small{
			Recall-precision tradeoff for YOLOv3 detector trained on DETRAC.
			The blue row in the table corresponds to the intersection of recall and precision curves and represents the best detector.
		}
	}
	\label{fig-yolo_rec_prec}
\end{figure}

The idea here was to validate the extent to which being able to train the detector to adapt to MDP would improve performance.
This can be done by simulating a recall-precision tradeoff for both the detector and MDP and then comparing the performance difference between matched and mismatched combinations of the two.
I simulated detection tradeoff by varying the confidence threshold of a YOLOv3 \cite{Redmon2018_yolov3} detector trained on DETRAC (Fig. \ref{fig-yolo_rec_prec}).
I then trained a new MDP model using each set of detections thus obtained to simulate tracking tradeoff
Finally, I performed three-way testing by pairing each tracker with its own detector, all trackers with the best detector, and the best tracker with all the detectors.
However, the results of these tests (Fig. \ref{fig-yolo_sweep_results}) showed that MDP learns very little from the detector so that the expedient of training models on detections with different recall-precision characteristics cannot be used to simulate trackers with similar tradeoffs.

\subsection{Dummy Policies}
\label{deep_mdp-dummy}
\begin{figure}[t]
	\centering
	\includegraphics[width=\textwidth]{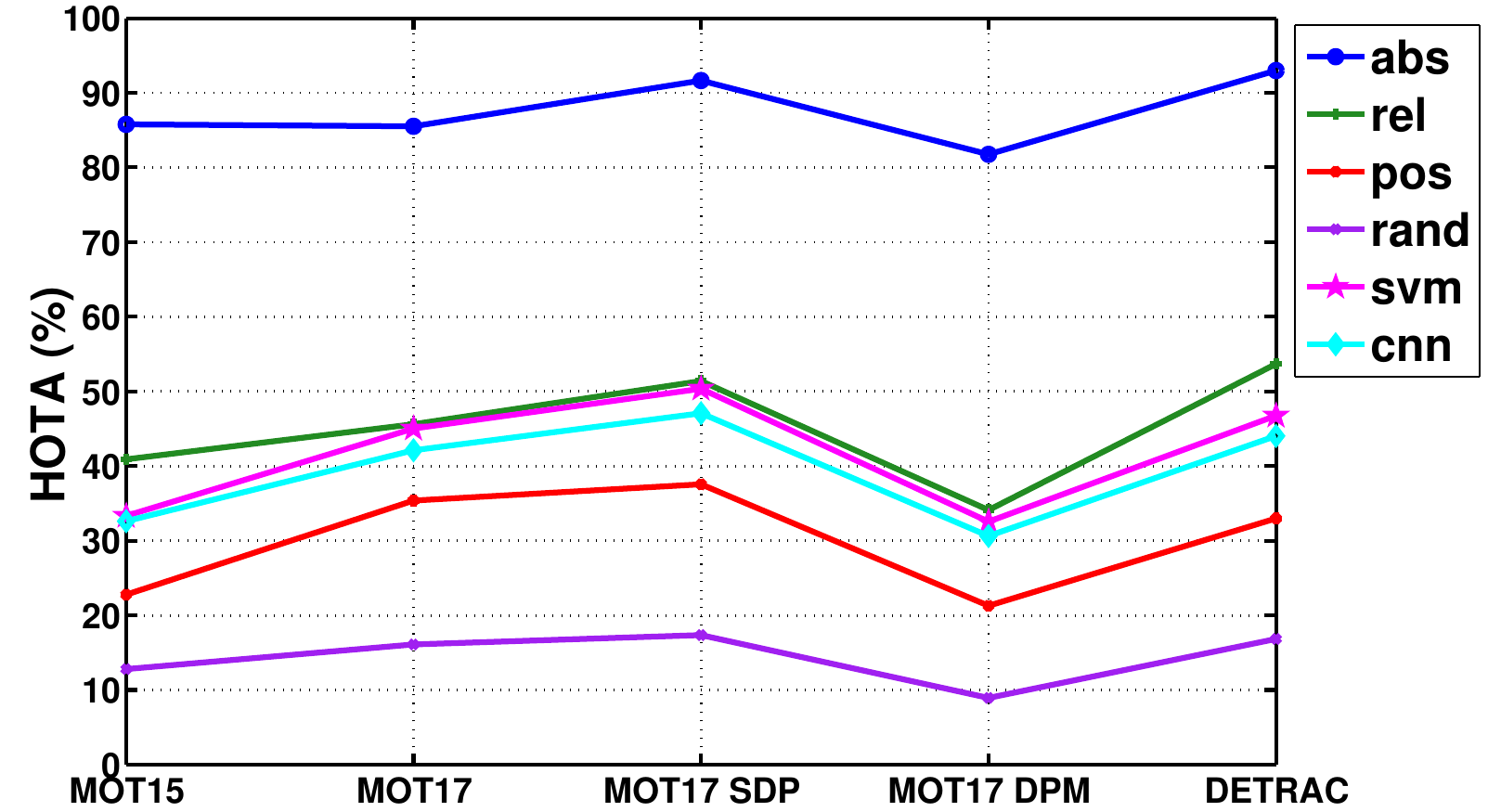}
	\caption{
		\small{
			Tracking performance in terms of HOTA for all dummy policies in addition to SVM and CNN on MOT2015, MOT2017 and DETRAC datasets.
			abs: absolute oracle, rel: relative oracle, pos: positive classifier, rand: random classifier.
			MOT17, MOT17 SDP and MOT17 DPM respectively refer to results obtained using MOT2017 public detections from Faster-RCNN \cite{Ren2015_faster_rcnn}, SDP \cite{Yang2016_sdp} and DPM \cite{Girshick2015_dpm} detectors.		 
		}
	}
	\label{fig-hota_dummy}
\end{figure}

In order to evaluate the impact of learning better policies on the overall MOT performance, I tested four dummy policies. 
Two of these are oracles that take the best possible decision given the GT:
\begin{itemize}[left=0pt,topsep=0pt,noitemsep,label=\textendash]
	\item Relative Oracle: This corresponds to how the policies are actually trained and its decisions depend on the detections, tracking result and other policies in addition to the GT.
	This represents an upper bound for the performance of real policies.	
	\item Absolute Oracle: This is an idealized policy for which no
	real policy can be trained
	%	training regime exists
	since it makes decisions using only the GT while ignoring all other cues.
	%	including the detections.
\end{itemize}
The remaining two policies make decisions independently of any external cues:
\begin{itemize}[left=0pt,topsep=0pt,noitemsep,label=\textendash]
	\item Positive Classifier: 	This classifies all samples as positive which results in a simplified policy-free  MDP where state transitions are governed entirely by presence or absence of detections since \textit{active}  adds all unassociated detections as new targets while \textit{tracked} and \textit{lost} always transition to \textit{tracked} unless a corresponding detection is missing.		
	\item Random Classifier: This classfies each sample randomly
	%	so that roughly half are classified as positive and negative.
\end{itemize}
As shown in Fig. \ref{fig-hota_dummy}, relative oracle does indeed perform very similarly to LK and CNN and much worse than absolute oracle which confirms my supposition that training better policies with deep learning is not going to significantly improve the overall tracking performance.
I also tried replacing LK with an oracle tracker (similar to oracle policies but for tracking) but that did not yield any useful insights.
%\subsubsection{GT Tracker}
%\subsubsection{Decision Statistics}

\subsection{Heuristics}
I tried lots of ad-hoc tricks to improve performance, some of which are listed below:
\begin{itemize}[left=0pt,topsep=0pt,noitemsep,label=\textendash]
	\item adding the predicted box from \textit{tracked} as a pseudo-detection when attempting to reconnect a recently \textit{lost} target
	\item removing excessive \textit{lost} targets by thresholding on the ratio of \textit{lost} to \textit{tracked} targets
	\item improved association method to generate classification GT by matching a detection-GT pair only if the former is the maximum-IOU detection for the latter and the latter is also the maximum-IOU GT for the former
	\item pre-training \textit{active} before running IBT iterations for \textit{tracked} and \textit{lost}
	\item ignoring detections in \textit{tracked} to make policy decision purely with the classifier
	%\item retraining \textit{active} for each sequence during inference
\end{itemize}

%\section{Results}
%\label{app-diff_tracking-evaluation}
%This section provides
%quantitative results for a small selection of my experiments with deep MDP along with some diagnostic data.
%Most of these results were generated using different subsets of the DETRAC dataset but similar relative performance holds for other datasets too.
%Discussions about these results have been relegated to the respective captions.

%\section{Results}
%\label{app-diff_tracking-results}
%This section presents the results from some of my experiments with deep MDP including diagnostics.

\section{Conclusions}
\label{sec_conclusions}
This paper presented a modular library for MOT that was designed to create an end-to-end differentiable pipeline composed of heterogenous and replaceable elements.
This goal ultimately proved to be infeasible due to inherent limitations in the MDP framework But the code base that was generated in the process should still be useful to researchers working in this field.
The negative results presented in this paper might also help others working on related ideas to avoid wasting time on similar experiments.

\appendices
\section{Differentiable approximations to MOT components}
\label{review-approx_diff}
This section includes trackers that approximate a specific part of the MOT pipeline by a differentiable alternative which might in future be useful in constructing a completely differentiable pipeline.
All three methods included here have proposed approximations for the association step that is typically performed using the Hungarian algorithm \cite{Kuhn1955_hungarian} with either handcrafted or learned costs.
To the best of my knowledge, this is the only MOT component for which differentiable approximations have so far been proposed.

DAN \cite{Sun2018_dan,Sun2021_dan_journal} has introduced one such approximation where the association matrix is directly predicted by limiting the maximum number of detectable objects in each frame to 80 so that the corresponding association matrix has a fixed size of 80 $\times$ 80.
Dummy rows and columns containing all zeros are used to fill up the matrix in cases of fewer objects.
Two different association matrices of size 81 $\times$ 80 and 80 $\times$ 81 are then constructed by respectively adding an extra row and column to the base matrix
These account for objects that disappear or appear in the second frame and carry out both forward and backward association.
Though the network that predicts the association matrix using a pairs of frames is trained on the binary association matrices constructed from GT, its output is not directly used to associate objects in the two frames.
Instead, heuristics are used to accumulate information from a certain number of past frames to generate a cost matrix
for Hungarian association.

%Instead of associating objects between only pairs of frames,
FAMNet \cite{Chu19_FAMNet} takes a more ambitious approach to perform global association over an entire batch of frames
rather than only frame-pairs
through a modified version of the rank 1 tensor approximation (R1TA) framework \cite{Shi2019_R1TA}.
This is implemented by a multi-dimensional assignment (MDA) subnet
%of the overall FAMNet architecture
which takes as input object similarities generated by the Siamese-style affinity subnet
which in turn takes object-level image features extracted by the feature subnet.
Though FAMNet claims to perform global multi-frame association, the basic idea of the R1TA itself involves decomposing the global assignment into a set of local assignments where only pairs of consecutive frames are considered.
Practical gains from associating over more than two frames thus remain dubious.
In addition, the paper only considers batches of 3 frames which are hardly any improvement over frame pairs anyway.
Finally, the process of generating association GT is full of heuristics, as is the formulation of the R1TA power iteration layer in the MDA subnet that has very suboptimal theoretical guarantees.
%and uses L1 normalization to approximately satisfy some of the original constraints

While the above methods attempt to approximate the binary association matrix that the Hungarian algorithm produces as output, DeepMOT \cite{Xu20_diff_motp} has proposed a Deep Hungarian Net (DHN) module to directly approximate the algorithm itself.
DHN uses two bidirectional RNNs to produce a soft approximation to the detection-to-GT assignments.
This in turn is used to propose differentiable proxies to the standard MOT metrics of MOTA and MOTP \cite{Bernardin2008_mota} so that the tracker can be optimized directly in terms of these approximate metrics.
Unlike DAN \cite{Sun2018_dan,Sun2021_dan_journal}, DHN is designed to handle variable sized assignment matrices.
%Assuming that there are $N$ detections and $M$ GT boxes, DHN takes as input an $N \times M$ distance matrix where the distance between each pair of boxes is the average of 1 - IOU and normalized Euclidean distance to avoid the problem IOU discontinuity at 1 which can be seen as an alternative to soft IOU \cite{Menshov2019_sort_soft_iou}.
It takes as input a matrix containing pairwise distances between boxes computed as the average of IOU and normalized Euclidean distances.
%This matrix is first subjected to row-wise flattening and passed through the first RNN to obtain a  $N \times M \times 2h$ output where $ h $ is the number of hidden units.
%This in turn is subjected to a column-wise flattening and passed through the second RNN to obtain a same size output which is converted to a final $N \times M$ assignment matrix through some fully connected layers.
This matrix is sequentially flattened row-wise and column-wise
before being passed through each of the two RNNs
to mimic the similar flattening in the Hungarian algorithm.
A row and a column containing 0.5 are appended to the soft assignment matrix generated by the DHN
to construct two additional matrices which are in turn subjected to column-wise and row-wise softmax respectively followed by summing the extra row and column to finally obtain soft approximations to FP and FN.
An approximation to IDS is likewise computed as the L1 norm of the element-wise product between the row-appended matrix above  and a binary matrix showing the locations of true positives.
The approximate FP, FN and IDS are then used to compute approximate MOTA and MOTP whose weighted average becomes the loss for training the DHN.

{\small
	\bibliographystyle{ieee}
	\bibliography{references}
}
\end{document}